\documentclass[runningheads]{llncs}

\usepackage{eccv}

\usepackage{eccvabbrv}

\usepackage{graphicx}
\usepackage{booktabs}
\usepackage{makecell}
\usepackage{pifont}
\newcommand{\blackcheck}{\ding{51}}
\newcommand{\blackcross}{\ding{55}}
\usepackage{multirow}

\usepackage{tcolorbox}
\tcbuselibrary{breakable,skins}
\usepackage{listings}
\usepackage[table]{xcolor}
\usepackage{multirow}
\usepackage{xcolor}

\lstdefinestyle{jsonstyle}{
  basicstyle=\ttfamily\small,
  columns=fullflexible,
  frame=none,
  xleftmargin=0pt,
  showstringspaces=false,
  breaklines=true,
  aboveskip=0pt,
  belowskip=0pt,
  escapeinside={(*@}{@*)}
}

\newcommand{\Human}{\textcolor{black}{X}\textsubscript{\textcolor{red}{\small human}}:}
\newcommand{\GPT}{\textcolor{black}{X}\textsubscript{\textcolor{red}{\small gpt}}:}
\newcommand{\imgtok}{\textcolor{green!60!black}{\mbox{\ttfamily\footnotesize<image>}}}
\newcommand{\starimgtok}{
\textcolor{green!60!black}{\mbox{\ttfamily\footnotesize<image>}}%
\textsuperscript{\scriptsize\textcolor{green!60!black}{$\star$}}%
\textsubscript{\tiny\textcolor{green!60!black}{(Frame-2)}}
}
\newcommand{\markedimgtok}{\textcolor{red!70!black}{\mbox{\ttfamily\footnotesize<marked\_image>}}}
\newcommand{\tasktitle}[1]{\noindent\textcolor{blue!70!black}{\bfseries\itshape #1}\par}
\newcommand{\imgdots}{\textcolor{green!60!black}{$\dots$}}

\usepackage{hyperref}

\usepackage{orcidlink}

\providecommand{\thetitle}{}
\let\oldtitle\title
\renewcommand{\title}[1]{\oldtitle{#1}\renewcommand{\thetitle}{#1}}
\newcommand{\maketitlesupplementary}{
    \newpage
    \begin{center}
        \Large
        \textbf{\thetitle}\\[0.5em] 
        Supplementary Material\\[1.0em]
    \end{center}
}

\usepackage[accsupp]{axessibility}  

\usepackage{hyperref}

\usepackage{orcidlink}

\begin{document}

\title{GAP-MLLM: Geometry-Aligned Pre-training for Activating 3D Spatial Perception in Multimodal Large Language Models} 

\titlerunning{GAP-MLLM}
\authorrunning{J.~Zhang et al.}
\author{
Jiaxin Zhang\textsuperscript{1\#} \and Junjun Jiang\textsuperscript{1\dag} \and Haijie Li\textsuperscript{2} \and Youyu Chen\textsuperscript{1} \and Kui Jiang\textsuperscript{1} \and \\ Dave Zhenyu Chen\textsuperscript{3}
}
\institute{
Harbin Institute of Technology, China \and
School of Electronic and Computer Engineering, Peking University, China \and 
Huawei, China \\
\textsuperscript{\#}Intern at Huawei. \quad
\textsuperscript{\dag}Corresponding Author.  \\
\url{https://gapmllm.github.io/} \\
}

\maketitle

\vspace{-6mm}
\input\begin{figure*}
  \centering
  \includegraphics[width=1.0\linewidth]{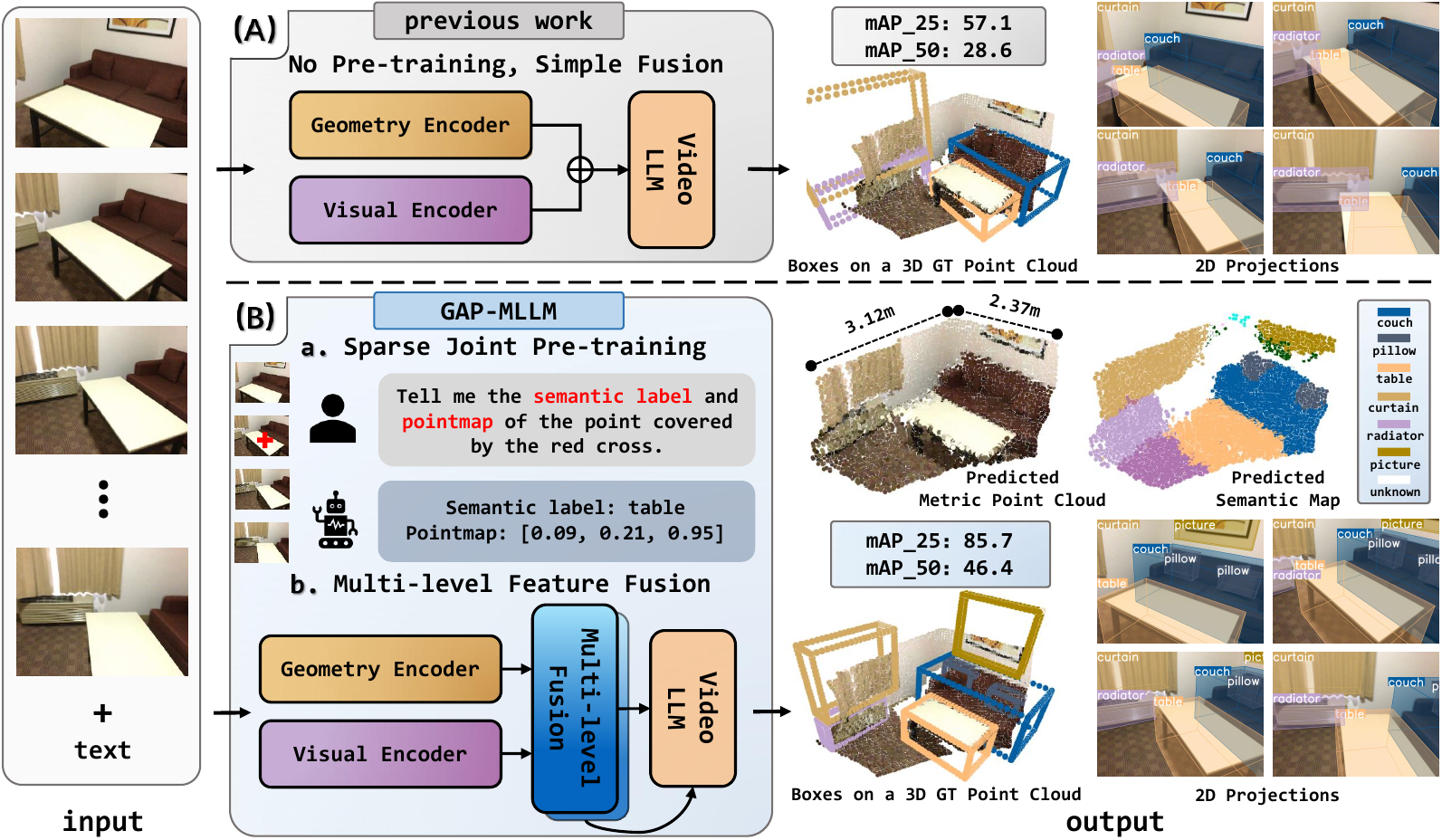}
  \vspace{-6mm}
	\captionof{figure}{Geometry-aligned pre-training significantly improves 3D perception in MLLMs. 
(A) Naive fusion without geometry-aware pre-training leads to limited geometric utilization and inaccurate 3D detection. 
(B) Our sparse geometry–semantics joint pre-training and multi-level fusion module progressively integrate geometric priors, activating structural perception and yielding substantial gains in 3D detection mAP.}
  \label{fig:example}
  \vspace{-6mm}
\end{figure*}

\begin{abstract}
Multimodal Large Language Models (MLLMs) demonstrate exceptional semantic reasoning but struggle with 3D spatial perception when restricted to pure RGB inputs. 
Despite leveraging implicit geometric priors from 3D reconstruction models, image-based methods still exhibit a notable performance gap compared to methods using explicit 3D data. 
We argue that this gap does not arise from insufficient geometric priors, but from a misalignment in the training paradigm: text-dominated fine-tuning fails to activate geometric representations within MLLMs.
Existing approaches typically resort to naive feature concatenation and optimize directly for downstream tasks without geometry-specific supervision, leading to suboptimal structural utilization.
To address this limitation, we propose \textbf{GAP-MLLM}, a \textbf{G}eometry-\textbf{A}ligned \textbf{P}re-training paradigm that explicitly activates structural perception before downstream adaptation. 
Specifically, we introduce a visual-prompted joint task that compels the MLLMs to predict sparse pointmaps alongside semantic labels, thereby enforcing geometric awareness.
Furthermore, we design a multi-level progressive fusion module with a token-level gating mechanism, enabling adaptive integration of geometric priors without suppressing semantic reasoning.
Extensive experiments demonstrate that GAP-MLLM significantly enhances geometric feature fusion and consistently enhances performance across 3D visual grounding, 3D dense captioning, and 3D video object detection tasks.
\end{abstract}
    
\section{Introduction}
\label{sec:intro}

Multimodal large language models (MLLMs), including vision-language models (VLMs)~\cite{bai2025qwen3vltechnicalreport,hurst2024gpt,li2024llava} and vision-language-action models~\cite{zitkovich2023rt,qu2025spatialvla,kim2024openvla}, have demonstrated strong cross-modal reasoning and semantic understanding capabilities. Extending these models to 3D spatial perception is a crucial step toward grounding MLLMs in the physical world~\cite{zheng2025multimodal,chen2025reasoning}, enabling structured scene understanding and object localization under natural language supervision.

Recent efforts~\cite{zheng2025learning, fan2025vlm} enhance 3D spatial perception under pure RGB input by incorporating geometric priors derived from feed-forward reconstruction models~\cite{wang2025continuous,wang2025vggt,wang2025pi}. These implicit geometric priors provide pixel-aligned structural cues without requiring explicit 3D inputs such as point clouds~\cite{SpatialLM, zheng2025video}, offering scalability and compatibility with vision-language pipelines. However, despite their promise, implicit prior-based methods consistently underperform in spatial perception compared to approaches operating on explicit 3D representations.

\begin{figure}[tb]
  \centering
  \includegraphics[width=1.0\linewidth]{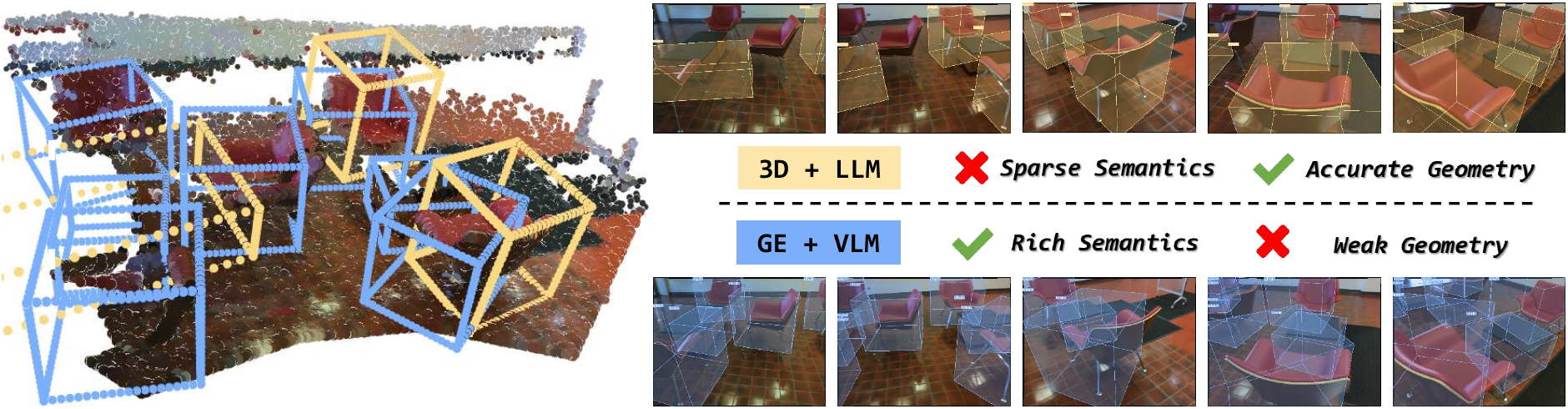}
	\captionof{figure}{\textbf{Geometry–semantics imbalance} in existing 3D perception paradigms. 
Point cloud-based methods ensure geometric accuracy but lack semantics, whereas image-based geometric encoders (GE) with VLMs retain semantics yet under-utilize geometry. 
As a result, both paradigms exhibit suboptimal performance in 3D perception.
}
  \label{fig:tensor1}
\end{figure}

We argue that this gap stems from a geometry–semantics imbalance in existing training paradigms rather than from insufficient geometric priors. 
As illustrated in Fig.~\ref{fig:tensor1}, point cloud-based methods~\cite{SpatialLM} possess strong geometric integrity but lack the high-level semantic reasoning of LLMs, whereas image-based MLLMs~\cite{zheng2025learning} enriched with geometric encoders preserve semantic prowess yet fail to fully exploit structural information for precise spatial inference.
Most prior works simply fuse geometric and visual features and fine-tune on text-prioritized downstream tasks without dedicated geometry-aware pre-training. 
Under this optimization regime, semantic supervision dominates the learning process, and geometric representations remain weakly activated, limiting their effective contribution to spatial perception and accurate localization.

To address this limitation,  we propose \textbf{GAP-MLLM}, a \textbf{G}eometry-\textbf{A}ligned \textbf{P}re-training paradigm that explicitly activates structural perception before task-specific learning. Rather than directly fine-tuning fusion modules on target tasks, we introduce a joint reconstruction-perception objective that aligns geometric priors with the inherent image-text representations of MLLMs across multiple feature levels.
Specifically, we design a vision-prompt-based joint task requiring MLLMs to predict both sparse pointmaps and semantic labels (see Fig.~\ref{fig:example}). 
This objective enforces structural reasoning as a primary capability rather than an auxiliary feature, preventing geometry from being suppressed by dominant language supervision. 
Furthermore, inspired by the hierarchical attention patterns of geometric encoders, we develop a multi-level progressive fusion module with a token-level gating mechanism. 
This design enables adaptive geometric integration from abstract global contexts to fine-grained local details, ensuring that geometry and semantics interact synergistically without mutual suppression.
Extensive experiments demonstrate that our geometry-aligned pre-training substantially enhances geometric feature utilization under sparse 3D supervision and consistently improves downstream performance on 3D visual grounding, 3D dense captioning, and 3D video object detection tasks.

\textit{Our contribution can be summarized as follows}:
\begin{enumerate}
    \item [1.] We propose a geometry-aligned pre-training paradigm that activates structural perception in image-only multimodal large language models with implicit geometric priors, moving beyond purely task-driven fine-tuning. 
    \item [2.] We develop a sparse geometry–semantics joint pre-training objective that aligns geometric priors with semantic representations through simultaneous prediction of 3D pointmaps and semantic labels. 
    \item [3.] We design a multi-level progressive fusion architecture with token-level gating, enabling hierarchical and adaptive integration of geometric priors from abstract to fine-grained representations across attention layers.
    \item [4.] We show through extensive experiments that GAP-MLLM effectively activates the spatial structure perception capabilities even with a small amount of sparse data, and as pre-trained weights, consistently improves performance across a diverse range of downstream 3D perception tasks.
\end{enumerate}

\section{Related Works}
\vspace{-2mm}

\subsubsection{Multimodal Large Language Models (MLLMs)}
Multimodal Large Language Models (MLLMs) have demonstrated strong semantic reasoning and visual–linguistic alignment capabilities across diverse 2D perception and reasoning tasks~\cite{radford2021learning,li2023blip,bai2025qwen2,bai2025qwen3vltechnicalreport,hurst2024gpt,li2024llava,team2024gemini,li2026causal}. 
Recent efforts~\cite{yang2025thinking,chen2025reasoning,wu2025spatial,feng2025video, ma2024whenllms, dwedari2023generating, gao2026map2thought, xu2025uniugg} extend these models to 3D spatial tasks such as grounding~\cite{scanrefer, chen2022d3net, chen2023unit3d, achlioptas2020referit3d}, captioning~\cite{chen2021scan2cap, chen2022d3net, chen2023unit3d}, and spatial object detection~\cite{cao2023coda, cao2025codav2, cao20243dgsdet, cao2026vggtdet, qi2019votenet, qi2020imvotenet}, aiming to bridge language understanding with real-world spatial structure. 
Some approaches incorporate explicit geometric inputs, including depth maps, point clouds, or BEV representations~\cite{zhu2024llava,qi2025gpt4scene,SpatialLM}, which improve spatial perception but rely on costly or scarce 3D annotations. 
Under pure RGB video settings, obtaining reliable 3D inputs and explicit geometry remains challenging, motivating the use of implicit geometric priors derived directly from image sequences.

\vspace{-3mm}

\subsubsection{Feed-Forward 3D Reconstruction}
Feed-forward 3D reconstruction provides an efficient mechanism for extracting geometric structure directly from RGB inputs.
Unlike traditional iterative pipelines~\cite{hartley2003multiple,schonberger2016structure}, recent approaches~\cite{wang2024dust3r,wang2025vggt,keetha2025mapanything,wang2025continuous,lin2025depth,wang2025pi} predict pixel-aligned 3D geometry end-to-end.
For example, DUSt3R~\cite{wang2024dust3r} estimates pointmaps without known camera parameters, while VGGT~\cite{wang2025vggt} and MapAnything~\cite{keetha2025mapanything} extend to multi-view and metric-scale reconstruction.
These models have also been applied to scene semantic segmentation~\cite{li2025iggt,koch2025unified,sheng2025spatialsplat}, efficient reconstruction~\cite{shen2025fastvggt,yuan2026infinitevggt}, and dynamic scene reconstruction~\cite{wang20254d}, demonstrating strong geometric expressiveness under pure RGB settings.
More importantly, feed-forward reconstruction provides scalable implicit geometric priors for multimodal systems. However, integrating reconstruction-derived geometry into MLLMs while preserving semantic reasoning remains challenging.
\vspace{-3mm}

\subsubsection{Image-based 3D Spatial Perception with MLLMs}
MLLM-powered image-based 3D perception methods aim to combine geometric understanding from image sequences with structured language reasoning.~\cite{SpatialLM,zheng2025learning,zheng2025video,fan2025vlm}. 
Existing approaches can be broadly divided into explicit-input and implicit-prior paradigms.
Explicit-input methods encode geometric representations into language-aligned tokens. 
SpatialLM~\cite{SpatialLM} reconstructs point clouds from images and converts them into structured tokens for spatial description, and Video-3D-LLM~\cite{zheng2025video} injects explicit 3D cues into visual tokens via positional encoding. 
While effective, these methods rely on high-quality 3D annotations.
Implicit-prior methods instead extract geometric cues from feed-forward reconstruction models and fuse them with visual features. 
VG-LLM~\cite{zheng2025learning} embeds a reconstruction encoder to inject geometric priors from video sequences, and VLM-3R~\cite{fan2025vlm} introduces spatial–visual–view fusion for spatial reasoning tasks. 
Although these approaches preserve strong semantic reasoning, most directly optimize fused representations on downstream tasks without geometry-aware pre-training. 
Under text-dominated supervision, geometric priors are often weakly activated, limiting structural perception and localization.
In contrast, our work focuses on a geometry-aligned pre-training paradigm that activates geometric priors before downstream adaptation, enabling more effective geometric utilization within MLLMs.

\begin{figure}[tb]
  \centering
  \includegraphics[width=1.0\linewidth]{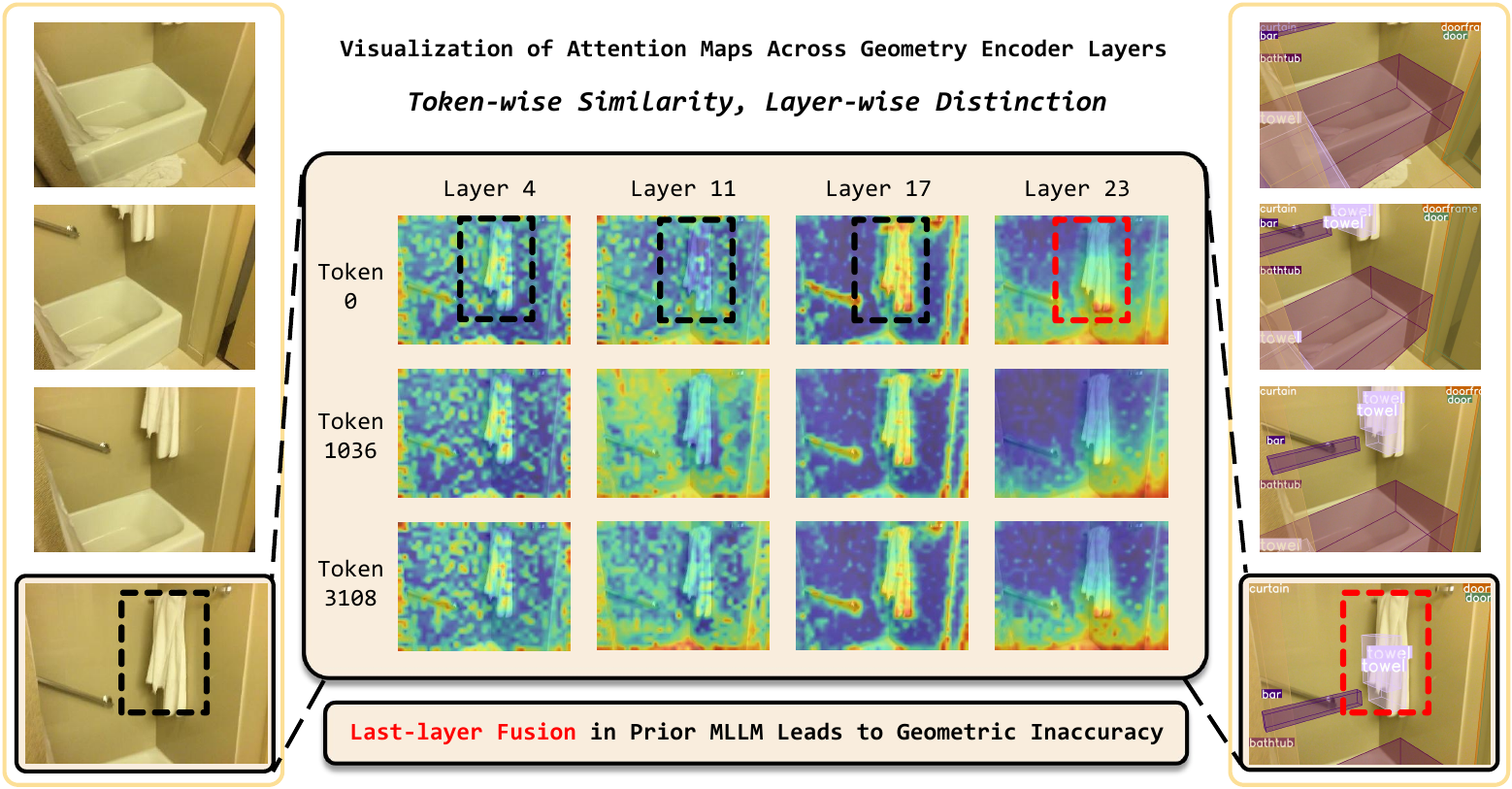}
  \vspace{-4mm}
	\captionof{figure}{\textbf{Failure Analysis.} Global attention maps of the geometric encoder are visualized using three representative tokens across layers. The same token exhibits distinct attention patterns at different layers. Prior method that relies on last-layer tokens as implicit geometric priors~\cite{zheng2025learning} leads to inaccurate 3D bounding box prediction.}
  \label{fig:tensor2}
\vspace{-2mm}
\end{figure}


\section{Method}
\vspace{-2mm}
We present \textbf{GAP-MLLM}, a geometry-aligned pre-training framework designed to activate structural perception in multimodal large language models under pure RGB input. 
The framework consists of a multi-level geometric–visual fusion architecture and a two-stage 3D perception training paradigm that jointly enhance metric-aware spatial perception. 
We first describe the architecture in Sec.~\ref{Architecture}, followed by the geometry-aligned training strategy in Sec.~\ref{Training}.

\vspace{-2mm}
\subsection{Architecture}
\label{Architecture}
\subsubsection{Visualization Analysis.}
Existing methods~\cite{zheng2025learning, fan2025vlm} enhance the spatial perception capabilities of MLLMs by extracting implicit geometric priors via geometric encoders with feed-forward reconstruction, but typically utilize only the last-layer features of geometric encoders as implicit priors. 
Recent works~\cite{shen2025fastvggt, yuan2026infinitevggt} have attempted to eliminate useless tokens by analyzing information redundancy in the intermediate feature space of feed-forward reconstruction. 
Different from these works that focus on token efficiency, we investigate how geometric representations at different layers affect downstream 3D perception when used as implicit priors.
As shown in Fig.~\ref{fig:tensor2}, we visualize the global attention maps of the geometric encoder across layers using representative tokens. 
The attention pattern of the same token varies significantly across layers, indicating that geometric cues are distributed hierarchically rather than concentrated in the last layer. 
When only the last-layer geometric tokens are injected to guide 3D bounding box prediction, the model exhibits inaccurate geometric structures caused by biased geometric attention. 
This observation suggests that intermediate-layer geometric representations are not sufficiently activated in existing pipelines, motivating the necessity of multi-level geometric feature fusion in GAP-MLLM.

\begin{figure}[tb]
  \centering
  \includegraphics[width=1.0\linewidth]{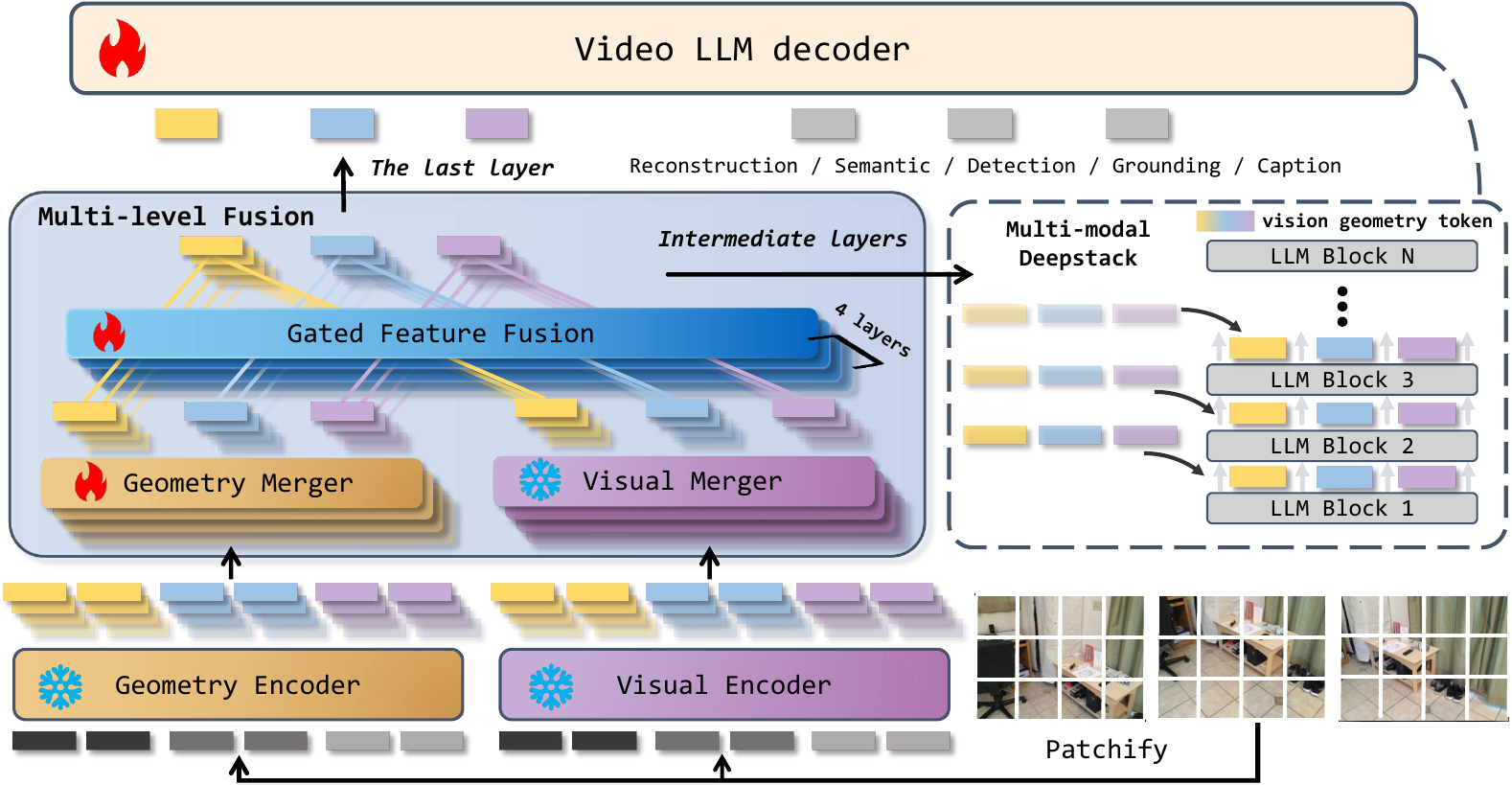}
  \vspace{-6mm}
    \captionof{figure}{\textbf{Network Architecture.} 
    An image sequence is processed by parallel geometric and visual branches to extract multi-level structural and semantic tokens. 
    After gated multi-level fusion, the final-layer fused tokens are aligned with task-related textual representations in the video LLM decoder, while selected intermediate-layer tokens are injected into early decoder blocks to preserve hierarchical geometric information.}
  \label{fig:pipeline}
\vspace{-4mm}
\end{figure}

\vspace{-2mm}

\subsubsection{Overview.}

Fig.~\ref{fig:pipeline} illustrates the overall architecture of GAP-MLLM. 
During both training and inference, the model takes an image sequence $\{I_i\}_{i=1}^{n}$ and a task-related natural language query $Q$ as input. 
The image sequence is processed by two parallel branches: a visual branch responsible for extracting semantic features and a geometric branch responsible for extracting structural representations derived from feed-forward reconstruction.
Specifically, for each image $I_i \in \mathbb{R}^{h \times w \times 3}$, the visual and geometric branches respectively generate multi-level tokens $\{\mathcal{T}_{i,j}^{V}\}_{j=1}^{L_V}$ and $\{\mathcal{T}_{i,j}^{G}\}_{j=1}^{L_G}$, where
\begin{equation}
\mathcal{T}_{i,j}^{V} \in \mathbb{R}^{\lfloor h/p_V \rfloor \times \lfloor w/p_V \rfloor \times c}, \quad
\mathcal{T}_{i,j}^{G} \in \mathbb{R}^{\lfloor h/p_G \rfloor \times \lfloor w/p_G \rfloor \times c}.
\end{equation}
Here, $p_V$ and $p_G$ denote the patch sizes of the two branches, and $L_V$ and $L_G$ represent the total number of layers. 
For simplicity and alignment between modalities, we set $p_V = p_G = p$ and $L_V = L_G = L$ in subsequent formulations.

These multi-level tokens are then passed into a gated fusion module, where geometric and visual representations are hierarchically integrated across layers. 
The fused tokens are subsequently aligned with textual representations in the MLLM decoder. 
The final-layer fused tokens serve as the primary input to the decoder, while selected intermediate-layer tokens are progressively injected during decoding to preserve multi-level structural information.
In this work, we adopt Qwen3-VL~\cite{bai2025qwen3vltechnicalreport} as the MLLM backbone and visual branch, and VGGT~\cite{wang2025vggt} as the geometric branch to extract implicit geometric priors.


\subsubsection{Multi-Level Feature Fusion.}
Existing implicit prior-based methods~\cite{zheng2025learning} typically fuse geometric and visual features via simple element-wise addition, relying only on last-layer geometric representations~\cite{fan2025vlm}. 
While effective, such static and single-layer fusion overlooks hierarchical geometric cues and may leave structural information insufficiently activated under dominant semantic supervision. 
To address this limitation, we introduce independent gating mechanisms across multiple layers for adaptive and hierarchical geometric integration.

The Multi-Level Fusion module takes multi-level visual tokens $\left\{ \mathcal{T}_{i,j}^V \right\}_{j=1}^{L}$ and geometric tokens $\left\{ \mathcal{T}_{i,j}^G \right\}_{j=1}^{L}$ as inputs. 
Note that the Qwen-VL series~\cite{bai2025qwen2,bai2025qwen3vltechnicalreport} compresses image tokens via token merging to reduce computational cost. 
To ensure spatial alignment between modalities, the geometric branch adopts an identical architectural design to the visual branch. 
Concretely, both branches group spatially adjacent $2 \times 2$ patches and pass them through a two-layer MLP to produce a single feature, yielding reduced visual tokens $\mathcal{T}_{i,j}^{V'} \in \mathbb{R}^{\lfloor \frac{h}{2p} \rfloor \times \lfloor \frac{w}{2p} \rfloor \times c}$ and geometric tokens $\mathcal{T}_{i,j}^{G'} \in \mathbb{R}^{\lfloor \frac{h}{2p} \rfloor \times \lfloor \frac{w}{2p} \rfloor \times c}$.

After token merging, we introduce an independent gating mechanism for each layer $j$ (where $j=1,2,\dots,L$) to dynamically balance geometric and visual contributions. 
The gating coefficient for layer $j$ is defined as:
\begin{equation}
g_{i,j} = \sigma\left( \text{MLP}\left( \left[ \mathcal{T}_{i,j}^{V'}, \mathcal{T}_{i,j}^{G'} \right] \right) \right),
\end{equation}
where $\sigma$ denotes the sigmoid activation function mapping values to $[0,1]$, $\text{MLP}$ represents a two-layer perceptron, and $\left[ \cdot, \cdot \right]$ indicates concatenation. 
The fused tokens $\left\{ \mathcal{T}_{i,j}^S \right\}_{j=1}^{L}$ are then computed as:
\begin{equation}
\mathcal{T}_{i,j}^S = g_{i,j} \odot \mathcal{T}_{i,j}^{V'} + (1 - g_{i,j}) \odot \mathcal{T}_{i,j}^{G'},
\end{equation}
where $\odot$ denotes element-wise multiplication. 
This per-layer token-level gating allows geometry–semantics weighting to vary across representation layers and spatial locations, ensuring that structural information remains sufficiently activated rather than uniformly blended.

To further enhance hierarchical structural preservation, we draw inspiration from DeepStack~\cite{meng2024deepstack} and Qwen3-VL~\cite{bai2025qwen3vltechnicalreport} to design a multi-modal DeepStack strategy. 
Specifically, from $\left\{ \mathcal{T}_{i,j}^S \right\}_{j=1}^{L}$, the final-layer tokens $\mathcal{T}_{i,L}^S$ are concatenated with textual embeddings as the primary input to the video LLM decoder. 
Meanwhile, selected intermediate-layer tokens $\mathcal{T}_{i,m}^S$ with $m \in \{L_1, L_2, L_3\}$ are injected into early decoder layers by direct addition to the corresponding hidden states. 
This hierarchical injection maintains geometric signals throughout decoding, preventing structural cues from being suppressed by dominant semantic representations and promoting sustained geometric activation.

\subsection{Training}
\label{Training}
\begin{figure}[tb]
  \centering
  \includegraphics[width=1.0\linewidth]{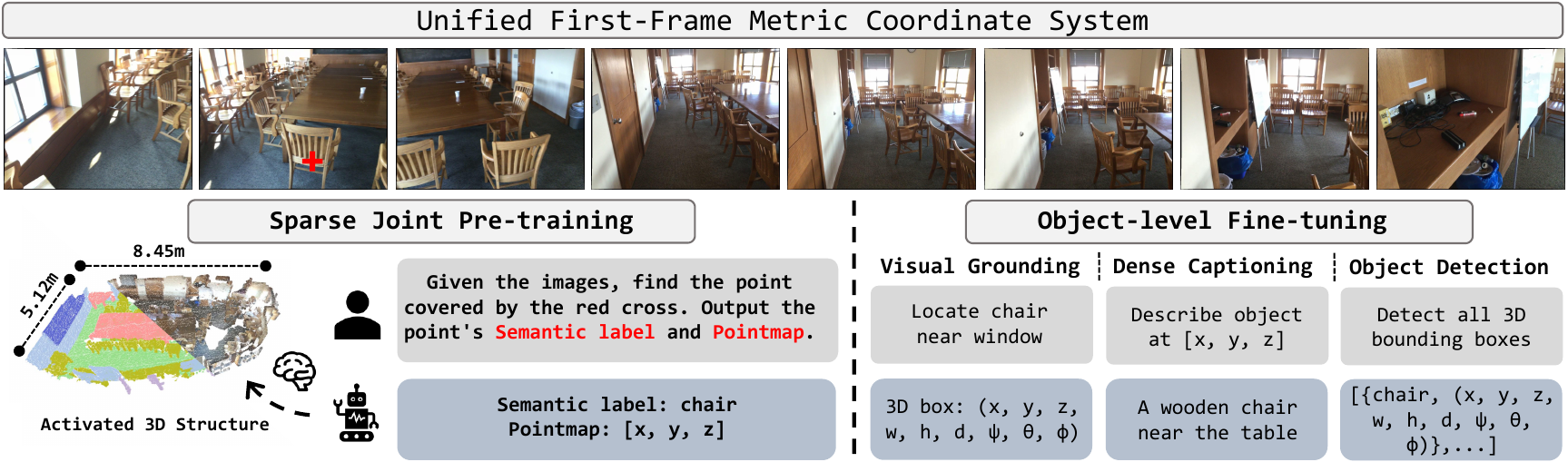}
  \vspace{-5mm}
    \captionof{figure}{\textbf{Training Strategy.} Sparse joint pre-training (left) activates 3D representations under a unified first-frame metric coordinate system.
The learned representations are then transferred to object-level fine-tuning (right) for downstream 3D perception.}
  \label{fig:training}
\vspace{-6mm}
\end{figure}

Recent works have proposed multiple object-level datasets for RGB-only 3D scene perception~\cite{zheng2025learning,yang2025visual}, covering tasks such as 3D visual grounding, dense captioning, and video object detection. 
However, directly fine-tuning MLLMs on object-level supervision remains insufficient to activate implicit geometric priors, as optimization is dominated by language-driven objectives. 
We argue that an explicit geometry-aligned pre-training stage is necessary to strengthen structural perception before downstream adaptation. 
Accordingly, we design a two-stage training paradigm, as illustrated in Fig.~\ref{fig:training}, consisting of sparse pixel-level geometry–semantics pre-training followed by object-level perception fine-tuning.

\vspace{-2mm}

\subsubsection{Consistent Metric Coordinate System.}
RGB-based 3D perception under implicit geometric priors lacks explicit constraints on the output space. 
Without a unified coordinate definition, predictions across tasks become inconsistent and undermine structural supervision.
To enforce geometry-aligned learning, we adopt a consistent metric coordinate system for all tasks. 
Following VGGT~\cite{wang2025vggt}, we use the first-frame coordinate system as the shared reference frame. 
All numerical outputs are represented in metric space and rounded to two decimal places, ensuring structural consistency across tasks.

\vspace{-3mm}

\subsubsection{Sparse Geometry-Semantics Joint Pre-training.}
Inspired by DepthLM~\cite{cai2025depthlm}, we observe that pure vision models still struggle with metric spatial perception. 
Even when equipped with implicit geometric priors, multi-view structural prediction remains weak under language-dominated supervision. 
Therefore, we introduce a sparse geometry–semantics joint pre-training objective to explicitly activate geometric representations under cross-modal guidance.

Given image sequences $\{I_1, \dots, I_n\}$, visual prompts, and a natural language query $Q$, the model predicts both the 3D coordinate $(x,y,z)$ of the prompted pixel and its semantic label $c$ under the unified first-frame coordinate system, as shown in Fig.~\ref{fig:training}. 
By jointly supervising structural coordinates and semantic categories, this objective aligns geometric priors with language representations and explicitly activates 3D structural perception during early optimization.

\vspace{-3mm}

\subsubsection{Object-level 3D Perception Fine-tuning.} After sparse geometry-semantics joint pre-training, we evaluate whether the activated structural representations transfer to complex object-level 3D perception tasks.

\noindent\textit{\textbf{3D Visual Grounding}}:
Following previous works~\cite{zhang2025flatland}, we formulate 3D visual grounding as a 3D video grounding problem that predicts the frame index where the target object appears and its 3D bounding box in that coordinate system. 
Unlike VG-LLM~\cite{zheng2025learning}, which performs grounding in a single feed-forward pass, we decompose it into a two-stage procedure. 
In the first stage, given an image sequence and a language query, the model predicts the target frame index $I_t$. 
In the second stage, we reorganize the sequence with $I_t$ as the first frame and predict the 3D bounding box $(x, y, z, w, h, d, \psi, \theta, \phi)$, where $(x, y, z)$ denotes the center coordinate, $(w, h, d)$ the object size, and $(\psi, \theta, \phi)$ the rotation angles.

\noindent\textit{\textbf{3D Dense Captioning}}:
Similar to visual grounding, we adopt a two-stage dense captioning approach~\cite{zhu2024llava}. 
Given a frame sequence and the spatial coordinate $(x, y, z)$ of an object center under the unified first-frame coordinate system, the model generates a descriptive caption conditioned on the corresponding 3D location. 
This formulation preserves coordinate consistency across tasks.

\noindent\textit{\textbf{3D Video Object Detection}}:
Following~\cite{zheng2025learning}, our 3D video object detection requires the model to output all objects appearing in a multi-frame sequence under the same first-frame coordinate system. 
The output is represented as $\{(b_i, c_i)\}$, where $b_i$ follows the same bounding box parameterization as in 3D visual grounding and $c_i$ denotes the object category.

\vspace{-2mm}
\section{Experiments}
\label{sec:exp}
\vspace{-2mm}

We evaluate \textbf{GAP-MLLM} on RGB-only 3D scene perception to validate our claim that geometry-aligned pre-training paradigm activates structural perception and improves implicit-prior utilization under sparse supervision. 
Sec.~\ref{Setting} describes datasets, baselines, metrics, and implementation details. 
Sec.~\ref{Evaluation} reports results on 3D visual grounding, dense captioning, and video object detection, and includes metric 3D reconstruction as a probe of metric-aware perception after sparse prompt-based pre-training. 
Sec.~\ref{Analysis} analyzes our sparse geometry-semantics pre-training and multi-level gated fusion.

\vspace{-3mm}

\subsection{Setting}
\label{Setting}

\subsubsection{Datasets and Baselines.}
We first conduct sparse pixel-level geometry semantics pre-training, and then perform multi-task fine-tuning on object-level RGB-only 3D perception datasets.

1. ~\textit{\textbf{Sparse Joint Pre-training}}:
We use EmbodiedScan~\cite{wang2024embodiedscan} and ScanNet~\cite{dai2017scannet}. 
To improve semantic label quality, we adopt frame-aligned 3D annotations from EmbodiedScan. 
Specifically, object centers are projected to 2D and then back-projected to metric 3D coordinates in the first-frame coordinate system using ScanNet depth maps and camera poses. 
Semantic labels are inherited from EmbodiedScan. 
In total, the pre-training set contains approximately 500K samples; however, since supervision is provided only at sparse prompted pixels, the effective pixel-level supervision is roughly equivalent to two 680$\times$480 images. 
Each training sample contains four consecutive frames sampled at 1 FPS, where a red cross marks the prompted pixel location.

2. ~\textit{\textbf{3D Visual Grounding}}:
We use ScanRefer~\cite{scanrefer} and formulate it as 3D video grounding: given a text query, the model predicts the target frame index and a 3D bounding box under that frame’s camera coordinates. 
Following VG-LLM~\cite{zheng2025learning}, EmbodiedScan~\cite{wang2024embodiedscan} annotations are used as ground-truth boxes.

3. ~\textit{\textbf{3D Dense Captioning}}:
We adopt Scan2Cap~\cite{chen2021scan2cap}. 
Mask3D proposals extracted by LEO~\cite{huang2023embodied} are used as input, and the model generates captions conditioned on object center coordinates under the first-frame coordinate system.

4. ~\textit{\textbf{3D Video Object Detection}}:
We use EmbodiedScan~\cite{wang2024embodiedscan} with frame-aligned dense 3D bounding boxes. 
Four consecutive frames are sampled at 1 FPS, and all annotations are transformed to the first-frame coordinate system~\cite{zheng2025learning}. 
The dataset is split into 958 training and 243 evaluation scenes.

{\it \textbf{Baselines.}}
We compare against both \emph{explicit 3D-input} methods and \emph{RGB-only} methods that rely on implicit geometric priors. 
For 3D visual grounding and captioning, we include task-specific models ScanRefer~\cite{scanrefer} and Scan2Cap~\cite{chen2021scan2cap}, as well as generalist 3D/LLM models Chat-3D v2~\cite{huang2024chat}, Grounded 3D-LLM~\cite{chen2024grounded}, LL3DA~\cite{chen2024ll3da}, LLaVA-3D~\cite{zhu2024llava}, Video-3D LLM~\cite{zheng2025video}, and VG-LLM~\cite{zheng2025learning}. 
For 3D video object detection, we compare with SpatialLM~\cite{SpatialLM} and VG-LLM~\cite{zheng2025learning}. 
For metric 3D reconstruction, we compare with CUT3R~\cite{wang2025continuous} and MapAnything~\cite{keetha2025mapanything}.

{\it \textbf{Metrics.}}
ScanRefer: Acc@0.25/0.5. 
Scan2Cap: CIDEr, BLEU-4, ROUGE on IoU $\ge$ 0.5. 
3D Video Detection: Precision/Recall/F1 at IoU 0.25 over 20 classes. 
Metric 3D Reconstruction: Accuracy/Completeness/Overall on 10 ScanNet~\cite{dai2017scannet} validation scenes under aligned and metric evaluation.

\vspace{-4mm}

\subsubsection{Implementation Details.}
Our main instantiation of GAP-MLLM is built upon Qwen3-VL-2B~\cite{bai2025qwen3vltechnicalreport}, with VGGT-1B~\cite{wang2025vggt} as the geometric encoder. 
To demonstrate architectural scalability, we also apply the same training paradigm and fusion design to the VG-LLM setting~\cite{zheng2025learning}.
Both sparse joint pre-training and mixed-task fine-tuning are conducted for one epoch using Adam, with a warm-up ratio of 0.03 and a peak learning rate of 1e-5. 
The batch size is 32 for joint pre-training and 16 for mixed-task fine-tuning. 
The visual and geometric encoders each contain $L=24$ layers. For multi-level fusion, we select intermediate layers $L_1=5$, $L_2=11$, and $L_3=17$, while the final layer $L$ is used as the primary decoder input.
We freeze the visual encoder of the MLLM and the geometric encoder to provide stable multimodal token representations, and optimize the MLLM backbone (LLM) together with the multi-level fusion module.

\vspace{-2.5mm}

\subsection{Evaluations}
\label{Evaluation}

\subsubsection{Results on 3D Visual Grounding.}

\begin{figure}[tb]
  \centering
  \includegraphics[width=1.0\linewidth]{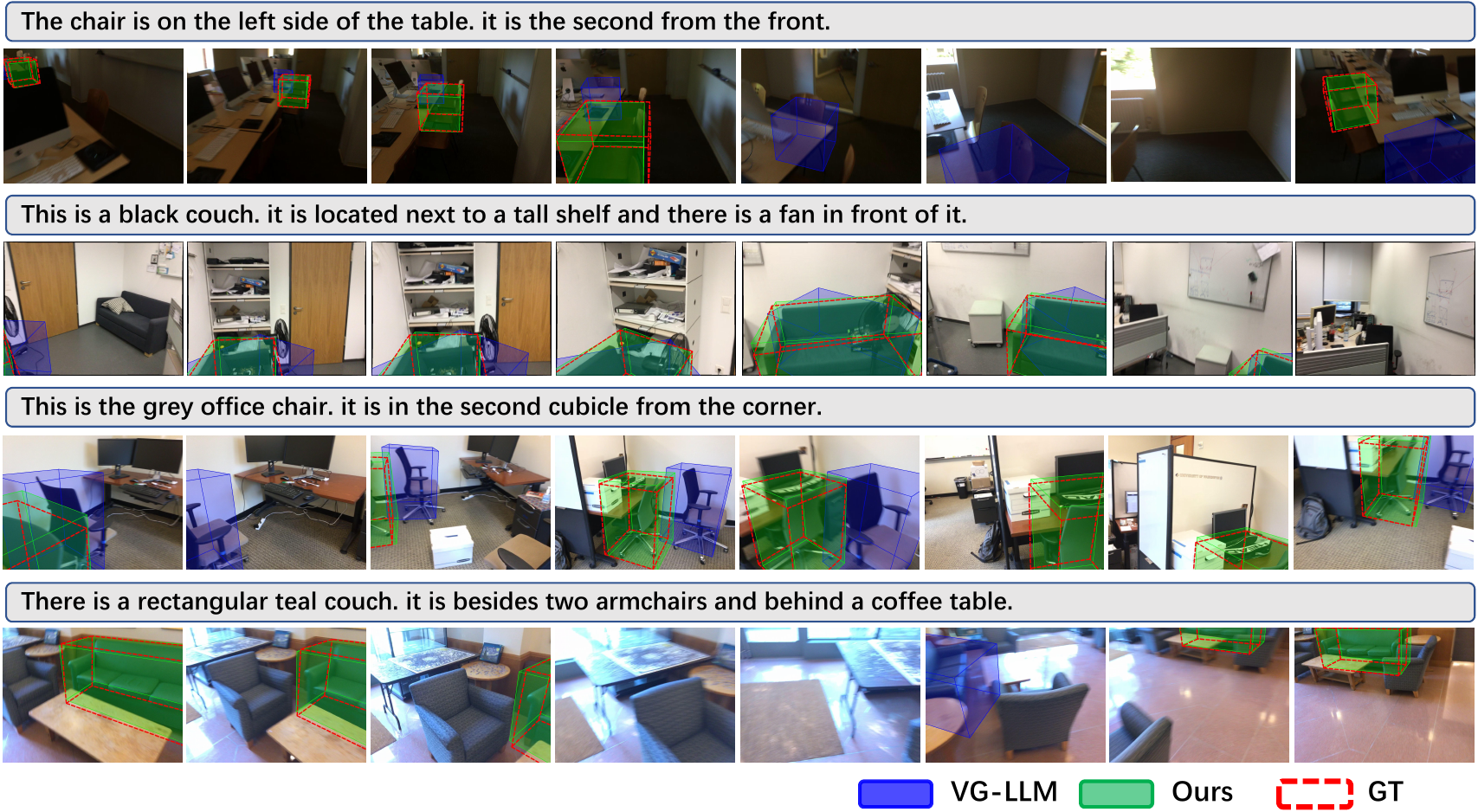}
  \vspace{-5mm}
	\captionof{figure}{\textbf{Qualitative comparison on 3D visual grounding.} 
We compare VG-LLM and GAP-MLLM against ground truth (GT). 
Our predictions exhibit tighter geometric alignment and more accurate box fitting under the same textual descriptions.}
  \label{fig:v1}
\vspace{-6mm}
\end{figure}

As shown in Tab.~\ref{tab:scanrefer}, GAP-MLLM significantly outperforms prior RGB-only methods that rely on implicit geometric priors. Compared with VG-LLM-4B~\cite{zheng2025learning}, our 3B model improves Acc@0.25 and Acc@0.5 by over 11 points while using fewer parameters, achieving comparable Acc@0.25 to 3D-input methods. These results indicate that geometry-aligned pre-training effectively activates structural perception within MLLMs.
Fig.~\ref{fig:v1} further illustrates that our predictions exhibit tighter geometric alignment and more accurate box structures. Even when both models identify the same object, our bounding boxes better conform to the underlying spatial geometry.

\begin{table*}[htbp]

\begin{minipage}[t]{0.45\linewidth}
\centering
\caption{\textbf{ScanRefer Results.} Ours surpasses prior RGB-only methods and approaches explicit 3D models.}
\vspace{-3mm}
\label{tab:scanrefer}
\resizebox{\linewidth}{!}{
\begin{tabular}{lc|cc}
\toprule
  {\textbf{Model}} & \makecell[c]{\textbf{3D Input}} & \textbf{Acc@0.25} & \textbf{Acc@0.5} \\
\midrule

ScanRefer~\cite{scanrefer} & \blackcheck & 37.3 & 24.3 \\
3D-LLM~\cite{hong20233d} & \blackcheck & 30.3 & - \\
Chat-3D v2~\cite{huang2024chat} & \blackcheck & 35.9 & 30.4 \\
Grounded 3D-LLM~\cite{chen2024grounded} & \blackcheck & 47.9 & 44.1 \\
ChatScene~\cite{huang2024chat} & \blackcheck & 55.5 & 50.2 \\
LLaVA-3D~\cite{zhu2024llava} & \blackcheck & 54.1 & 42.4 \\
Video-3D LLM~\cite{zheng2025video} & \blackcheck & \textbf{58.1} & \textbf{51.7} \\
\midrule
SPAR~\cite{zhang2025flatland} & \blackcross & 31.9 & 12.4 \\
VG-LLM-4B~\cite{zheng2025learning} & \blackcross & 36.4 & 11.8\\
VG-LLM-4B (w/ GAP) & \blackcross & 49.7  & 23.2 \\
GAP-MLLM-3B (Ours) & \blackcross & \textbf{53.1} & \textbf{26.0} \\
\bottomrule
\end{tabular}
}
\vfill
\end{minipage}
\hfill
\begin{minipage}[t]{0.52\linewidth}
\centering
\caption{\textbf{Scan2Cap Results.} GAP-MLLM achieves the best overall performance among both 3D-input and RGB-only methods.}
\vspace{-3mm}
\label{tab:scan2cap}
\resizebox{\linewidth}{!}{
\begin{tabular}{lc|ccc}
\toprule
{\textbf{Model}} & 
\makecell[c]{\textbf{3D Input}} & 
\textbf{C@0.5$\uparrow$} & 
\textbf{B-4@0.5$\uparrow$} & 
\textbf{R@0.5$\uparrow$} \\
\hline

Scan2Cap~\cite{chen2021scan2cap} & \blackcheck & 39.1 & 23.3 & 44.8 \\
LL3DA~\cite{chen2024ll3da} & \blackcheck & 65.2 & 36.8 & 55.0 \\
Chat-3D-v2~\cite{huang2024chat} & \blackcheck & 63.9 & 31.8 & - \\
Grounded 3D-LLM~\cite{chen2024grounded} & \blackcheck & 70.2 & 35.0 & - \\
LEO~\cite{huang2023embodied} & \blackcheck & 72.4 & 38.2 & 58.1 \\
Chat-Scene~\cite{huang2024chat} & \blackcheck & 77.1 & 36.5 & - \\
LLaVA-3D~\cite{zhu2024llava} & \blackcheck & 79.2 & \textbf{41.1} & \textbf{63.4} \\
Video-3D LLM~\cite{zheng2025video} & \blackcheck & \textbf{80.0} & 40.2 & 61.7 \\
\midrule
VG-LLM-4B~\cite{zheng2025learning} & \blackcross & 78.6 & 40.9 & 62.4 \\
VG-LLM (w/ GAP) & \blackcross & 80.6 & 41.1 & 62.8 \\
GAP-MLLM-3B (Ours) & \blackcross & \textbf{84.7} & \textbf{42.1} & \textbf{63.1} \\
\bottomrule
\end{tabular}
}
\vfill
\end{minipage}

\vspace{-2mm}
\end{table*}


\subsubsection{Results on 3D Dense Captioning.}
Tab.~\ref{tab:scan2cap} shows that GAP-MLLM achieves the best performance among RGB-only approaches and surpasses several explicit 3D-input methods. Our model obtains 84.7 CIDEr and 42.1 BLEU-4 at IoU 0.5, demonstrating that geometry-aligned pre-training improves coordinate-conditioned grounding and description in image sequences.

\begin{table*}[htbp]
\scriptsize
\centering
\caption{\textbf{3D Video Object Detection Results.} GAP-MLLM achieves the highest precision, recall, and F1 under both 4-frame and 6-frame settings.}
\vspace{-3.5mm}
\resizebox{0.9\linewidth}{!}{
\setlength{\tabcolsep}{2mm}
\begin{tabular}{l c|ccc|ccc}
\toprule
\multirow{2}{*}{\textbf{Model}} 
& \multirow{2}{*}{\textbf{3D Input}}
& \multicolumn{3}{c|}{\textbf{4-Frame Setting}} 
& \multicolumn{3}{c}{\textbf{6-Frame Setting}} \\
\cmidrule{3-5}\cmidrule{6-8}
& ~ 
& P$_{25}$ & R$_{25}$ & F1$_{25}$ 
& P$_{25}$ & R$_{25}$ & F1$_{25}$ \\
\midrule

SpatialLM~\cite{SpatialLM} 
& \blackcheck
& 40.8 & 23.7 & 29.1 
& 42.1 & 26.1 & 31.1 \\

\midrule

VG-LLM-4B~\cite{zheng2025learning}
& \blackcross
& 41.7 & 35.7 & 38.2 
& 39.7 & 34.0 & 36.4 \\

VG-LLM-8B~\cite{zheng2025learning}  
& \blackcross
& 43.4 & 39.6 & 41.2 
& 43.5 & 38.7 & 40.8 \\

VG-LLM-4B (w/ GAP) 
& \blackcross
& 47.0 & 41.2 & 43.5 
& 48.8 & 40.2 & 43.6 \\

GAP-MLLM-3B (Ours)
& \blackcross
& \textbf{54.2} & \textbf{48.1} & \textbf{50.6} 
& \textbf{52.9} & \textbf{45.4} & \textbf{48.5} \\

\bottomrule
\end{tabular}
}
\label{tab:3d_det_exp}
\vspace{-3mm}
\end{table*}


\subsubsection{Results on 3D Video Object Detection.}

\begin{figure}[tb]
  \centering
  \includegraphics[width=1.0\linewidth]{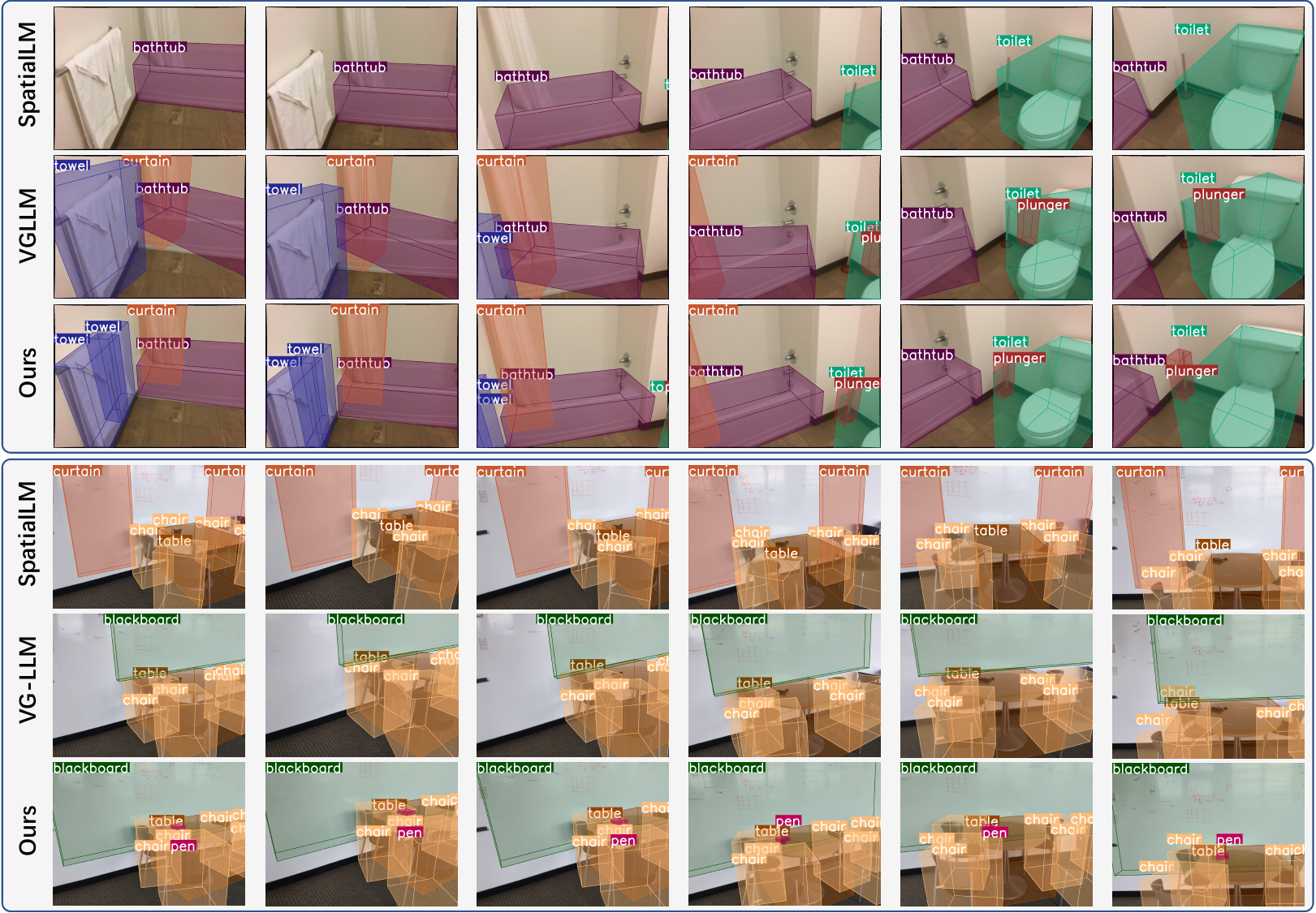}
  \vspace{-6mm}
	\captionof{figure}{\textbf{Qualitative comparison on 3D video object detection.} 
We compare SpatialLM~\cite{SpatialLM}, VG-LLM~\cite{zheng2025learning}, and GAP-MLLM. 
Our method produces more accurate and spatially consistent 3D bounding boxes across multi-frame inputs.}
  \label{fig:v2}
\vspace{-7mm}
\end{figure}

As shown in Tab.~\ref{tab:3d_det_exp} and Fig.~\ref{fig:v2}, SpatialLM~\cite{SpatialLM}, which relies on explicit 3D input, achieves reasonable precision but low recall, indicating sensitivity to incomplete geometry. RGB-based VG-LLM~\cite{zheng2025learning} improves recall through stronger semantic perception but remains limited in overall F1.
In contrast, GAP-MLLM consistently achieves the highest precision, recall, and F1 under both 4-frame and 6-frame settings. These results demonstrate that geometry-aligned pre-training and gated fusion enable more effective utilization of implicit geometric priors for accurate spatial localization.


\subsubsection{Results on Pointmaps and Semantic Labels.}

\begin{figure}[tb]
  \centering
  \includegraphics[width=1.0\linewidth]{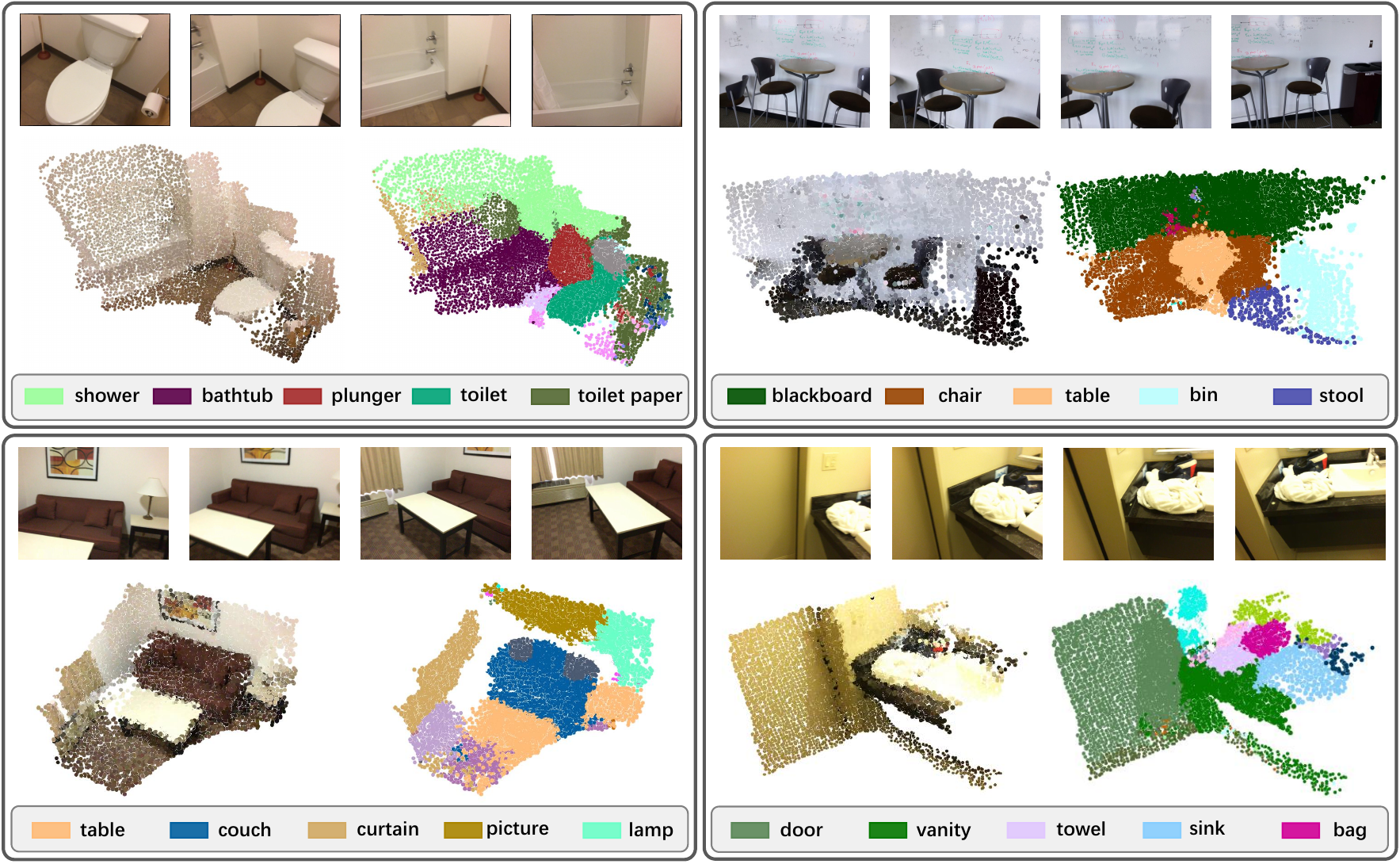}
  \vspace{-3mm}
\captionof{figure}{
\textbf{Qualitative results on pointmap and semantic prediction.} 
Despite being trained with sparse supervision, GAP-MLLM reconstructs metrically consistent 3D pointmaps and dense semantic regions.
White points denote the \emph{unknown} category.
}
  \label{fig:v3}
\vspace{-3mm}
\end{figure}

Our sparse joint pre-training enables the MLLM to generate metric-consistent point and semantic maps. As shown in Fig.~\ref{fig:v3}, the reconstructed geometry is structurally coherent and spatially accurate under metric supervision. The predicted semantic maps inherit the abstract semantic capacity of the MLLMs. Although boundaries are not sharp due to sparse supervision, the model reliably captures object extents and structure.

Tab.~\ref{tab:scannet_recon} reports aligned and metric evaluations on 10 random ScanNet~\cite{dai2017scannet} scenes. While CUT3R~\cite{wang2025continuous} achieves slightly better performance under Sim(3) aligned evaluation, GAP-MLLM performs best under metric evaluation, indicating stronger metric-aware reconstruction. Furthermore, semantic supervision yields improvements in metric consistency compared to pointmap only training, highlighting the role of semantic guidance in stabilizing metric structure.

\begin{table*}[htbp]
\centering
\caption{
\textbf{PointMap Estimation Results on ScanNet (meter).}
We report Accuracy, Completeness, and Overall under aligned evaluation (after Sim(3) alignment to GT) 
and metric evaluation (direct comparison with GT without alignment). 
}
\vspace{-2mm}
\resizebox{\linewidth}{!}{
\setlength{\tabcolsep}{1.5mm}
\renewcommand{\arraystretch}{1.2} 
\begin{tabular}{l|cccccc|cccccc}
\toprule
\multirow{3}{*}{\textbf{Model}} 
& \multicolumn{6}{c|}{\textbf{Aligned Evaluation}} 
& \multicolumn{6}{c}{\textbf{Metric Evaluation}} \\
\cmidrule(lr){2-7} \cmidrule(lr){8-13}
& \multicolumn{2}{c}{Acc. $\downarrow$} 
& \multicolumn{2}{c}{Comp. $\downarrow$} 
& \multicolumn{2}{c|}{Overall. $\downarrow$}
& \multicolumn{2}{c}{Acc. $\downarrow$} 
& \multicolumn{2}{c}{Comp. $\downarrow$} 
& \multicolumn{2}{c}{Overall. $\downarrow$} \\
\cmidrule(lr){2-3} \cmidrule(lr){4-5} \cmidrule(lr){6-7}
\cmidrule(lr){8-9} \cmidrule(lr){10-11} \cmidrule(lr){12-13}
& Mean & Med. 
& Mean & Med. 
& Mean & Med.
& Mean & Med. 
& Mean & Med. 
& Mean & Med. \\
\midrule

CUT3R~\cite{wang2025continuous} 
& 0.0302 & 0.0227 & 0.0273 & 0.0207 & 0.0287 & 0.0217
& 0.1199 & 0.0998 & 0.1045 & 0.0871 & 0.1122 & 0.0937 \\

Mapanything ~\cite{keetha2025mapanything}
& 0.0688 & 0.0397 & 0.0610 & 0.0480 & 0.0649 & 0.0438
& 0.3409 & 0.2721 & 0.2498 & 0.2265 & 0.2953 & 0.2493 \\

GAP-MLLM-3B (w/o semantic)
& 0.0424 & 0.0295 & 0.0326 & 0.0246 & 0.0375 & 0.0270
& 0.0691 & 0.0580 & 0.0570 & 0.0465 & 0.0630 & 0.0523 \\

GAP-MLLM-3B (Ours)
& 0.0421 & 0.0293 & 0.0327 & 0.0247 & 0.0374 & 0.0270
& 0.0675 & 0.0553 & 0.0557 & 0.0441 & 0.0616 & 0.0497 \\

\bottomrule
\end{tabular}
}
\label{tab:scannet_recon}
\end{table*}


\subsection{Analysis}
\label{Analysis}

\begin{table*}[htbp]
\scriptsize
\centering
\caption{
\textbf{Ablation on 3D video object detection.} Both sparse joint pre-training (A) and multi-level gated fusion (B) contribute complementary gains.}
\vspace{-2mm}
\setlength{\tabcolsep}{3mm}
\renewcommand{\arraystretch}{1.15}

\begin{tabular}{cc|ccc|ccc}
\toprule
\multirow{2}{*}{A} & \multirow{2}{*}{B}
& \multicolumn{3}{c|}{\makecell{\textbf{VG-LLM-4B~\cite{zheng2025learning}} \\ (Qwen2.5-VL + VGGT)}} 
& \multicolumn{3}{c}{\makecell{\textbf{GAP-MLLM-3B} \\ (Qwen3-VL + VGGT)}} \\
\cmidrule(lr){3-5} \cmidrule(lr){6-8}
& 
& P$_{25}$ & R$_{25}$ & F1$_{25}$ 
& P$_{25}$ & R$_{25}$ & F1$_{25}$ \\
\midrule

 &  
& 41.7 & 35.7 & 38.2 
& 48.8 & 42.1 & 44.7 \\

\checkmark &  
& 44.1 & 39.6 & 41.4 
& 52.7 & 45.9 & 48.7 \\

 & \checkmark 
& 43.3 & 36.9 & 39.4 
& 51.8 & 44.6 & 47.5 \\

\checkmark & \checkmark 
& \textbf{47.0} & \textbf{41.2} & \textbf{43.5} 
& \textbf{54.2} & \textbf{48.1} & \textbf{50.6} \\

\bottomrule
\end{tabular}

\label{tab1:ablation_detection}
\end{table*}

\subsubsection{Ablation Study on Core Contribution.}

Tab.~\ref{tab1:ablation_detection} shows that both sparse joint pre-training (A) and multi-level fusion (B) improve metrics in 3D video object detection. When combined, they achieve the best performance, demonstrating complementary effects.
Applying GAP to VG-LLM~\cite{zheng2025learning} yields consistent gains, confirming that our geometry-aligned design generalizes across architectures.


\subsubsection{Sparse Geometry-Semantics Joint Pre-training.}
To examine whether geometry aware pre-training enhances inherent spatial perception, we evaluate a pure Qwen3-VL~\cite{bai2025qwen3vltechnicalreport} backbone without geometric modules. As shown in Tab.~\ref{tab:pure_vlm}, sparse joint pre-training consistently improves all metrics, indicating strengthened geometric perception even without explicit 3D inputs or implicit priors.


We further compare pointmap-only and joint geometric–semantic supervision on VG-LLM~\cite{zheng2025learning}. As shown in Tab.~\ref{tab:joint_vs_pointmap}, joint supervision yields additional gains, suggesting that semantic guidance helps structure geometric representations. This trend aligns with the metric reconstruction results in Tab.~\ref{tab:scannet_recon}.

\begin{table*}[tb]
\scriptsize
\centering
\renewcommand{\arraystretch}{1.15}

\begin{minipage}[t]{0.46\linewidth}
\centering
\caption{\textbf{Effect of geometry-aware pre-training on pure VLM.} Pre-training enhances geometric perception, even without geometric encoder.}
\vspace{-1mm}
\begin{tabular}{lccc}
\toprule
\textbf{Model} & P$_{25}$ & R$_{25}$ & F1$_{25}$ \\
\midrule
Qwen3-VL & 43.0 & 37.3 & 39.7 \\
+ Joint Pre-training & \textbf{45.5} & \textbf{40.1} & \textbf{42.4} \\
\bottomrule
\end{tabular}
\label{tab:pure_vlm}
\end{minipage}
\hfill
\begin{minipage}[t]{0.50\linewidth}
\centering
\caption{\textbf{Effect of joint geometric-semantic pre-training.}
Joint supervision outperforms pointmap-only training in 3D perception.}
\vspace{-2mm}
\begin{tabular}{lccc}
\toprule
\textbf{Model} & P$_{25}$ & R$_{25}$ & F1$_{25}$ \\
\midrule
VG-LLM~\cite{zheng2025learning} & 41.7 & 35.7 & 38.2 \\
+ Pointmap & 43.1 & 38.3 & 40.1 \\
+ Pointmap + Semantic & \textbf{44.1} & \textbf{39.6} & \textbf{41.4} \\
\bottomrule
\end{tabular}
\label{tab:joint_vs_pointmap}
\end{minipage}
\end{table*}

\subsubsection{Ablation on Fusion Strategies.}
\begin{figure*}[htbp]
\centering
\begin{minipage}{0.40\linewidth}
\vspace{-3mm}
\centering
\captionof{table}{\textbf{Ablation on token-level fusion strategies.} Gated fusion achieves the best performance with higher flexibility.}
\vspace{2mm}
\scriptsize
\renewcommand{\arraystretch}{1.15}
\begin{tabular}{lccc}
\toprule
\textbf{Fusion Type} & Precision & Recall & F1 \\
\midrule
Add & 51.7 & 45.3 & 47.9 \\
Weighted & 51.4 & 45.5 & 47.8 \\
Cross-Attention & 52.7 & 46.3 & 49.1 \\
Gated (Ours) & \textbf{54.2} & \textbf{48.1} & \textbf{50.6} \\
\bottomrule
\end{tabular}
\label{tab:fusion_ablation}
\end{minipage}
\hfill
\begin{minipage}{0.56\linewidth}
\centering
\includegraphics[width=\linewidth]{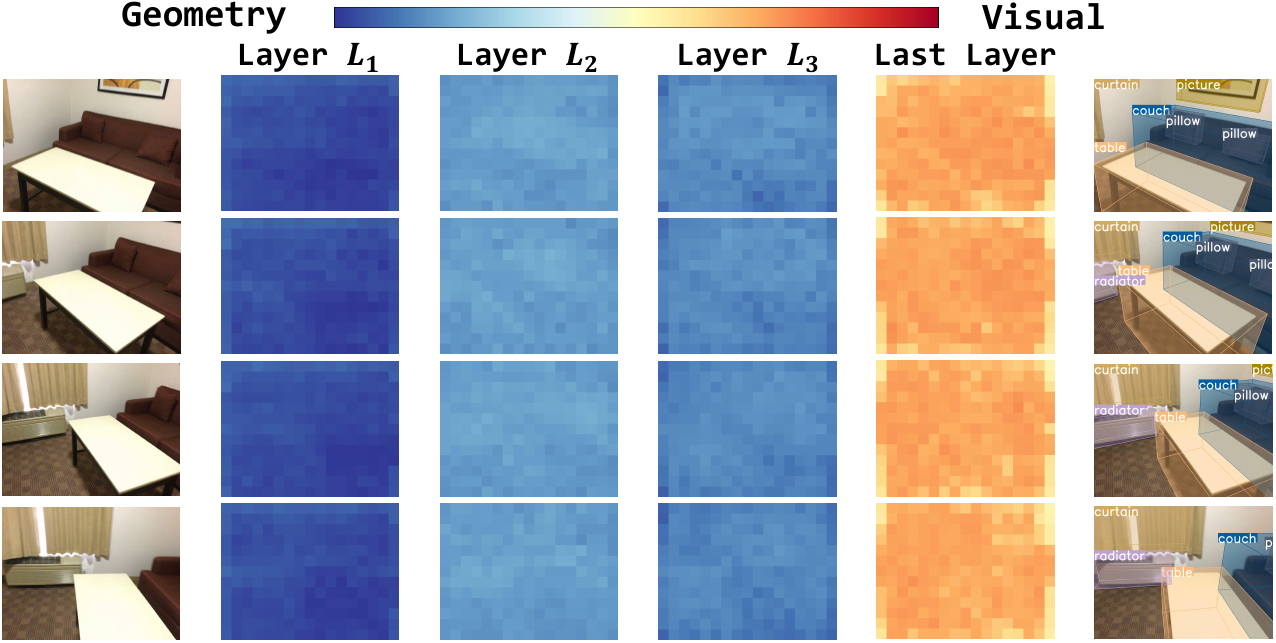}
\vspace{-5mm}
\captionof{figure}{\textbf{Visualization of Layer-wise gating weights.} Intermediate layers emphasize geometry, while later layers focus on semantics.}
\label{fig:v4}
\end{minipage}
\end{figure*}

Tab.~\ref{tab:fusion_ablation} compares different fusion strategies. Simple addition or weighted fusion provides limited gains, while cross-attention improves interaction. Our gated fusion achieves the best results, highlighting the importance of adaptive token-level integration.

\vspace{-4mm}
\vspace{-5mm}

Fig.~\ref{fig:v4} visualizes layer-wise gating weights. Intermediate layers emphasize geometric priors, while the final layer assigns higher weight to semantic features. This layer-dependent behavior confirms that geometry and semantics play complementary roles across stages, and gated fusion enables flexible balancing.

\section{Conclusion}
In this work, we revisit RGB-only 3D perception from a training perspective and argue that the performance gap of implicit geometric prior-based methods does not stem from insufficient geometry, but from the lack of effective activation within text-dominated fine-tuning paradigms. 
To address this issue, we propose \textbf{GAP-MLLM}, a geometry-aligned pre-training paradigm that explicitly activates structural perception before downstream adaptation. Through sparse geometry–semantics joint supervision and multi-level gated fusion, GAP-MLLM strengthens metric-aware spatial perception while preserving semantic capability. 
Extensive experiments across 3D visual grounding, 3D dense captioning, 3D video object detection, and metric 3D reconstruction demonstrate that geometry-aligned pre-training consistently improves the utilization of implicit geometric priors and generalizes across different MLLM backbones.
We hope this work provides a new perspective on training MLLMs for 3D spatial understanding and highlights the importance of activation-oriented pre-training for multimodal perception in the absence of explicit 3D inputs.

\noindent\textbf{Acknowledgments.} The research was supported by the National Natural Science Foundation of China (62471158, U23B2009).



%
%
\bibliographystyle{splncs04}
\bibliography{main}

@String(ECCV  = {Eur. Conf. Comput. Vis.})

@String(AAAI  = {AAAI})

@String(ECCV  = {ECCV})

@inproceedings{zitkovich2023rt,
  title={Rt-2: Vision-language-action models transfer web knowledge to robotic control},
  author={Zitkovich, Brianna and Yu, Tianhe and Xu, Sichun and Xu, Peng and Xiao, Ted and Xia, Fei and Wu, Jialin and Wohlhart, Paul and Welker, Stefan and Wahid, Ayzaan and others},
  booktitle={Conference on Robot Learning},
  pages={2165--2183},
  year={2023},
  organization={PMLR}
}

@article{kim2024openvla,
  title={Openvla: An open-source vision-language-action model},
  author={Kim, Moo Jin and Pertsch, Karl and Karamcheti, Siddharth and Xiao, Ted and Balakrishna, Ashwin and Nair, Suraj and Rafailov, Rafael and Foster, Ethan and Lam, Grace and Sanketi, Pannag and others},
  journal={arXiv preprint arXiv:2406.09246},
  year={2024}
}

@article{qu2025spatialvla,
  title={Spatialvla: Exploring spatial representations for visual-language-action model},
  author={Qu, Delin and Song, Haoming and Chen, Qizhi and Yao, Yuanqi and Ye, Xinyi and Ding, Yan and Wang, Zhigang and Gu, JiaYuan and Zhao, Bin and Wang, Dong and others},
  journal={arXiv preprint arXiv:2501.15830},
  year={2025}
}

@inproceedings{radford2021learning,
  title={Learning transferable visual models from natural language supervision},
  author={Radford, Alec and Kim, Jong Wook and Hallacy, Chris and Ramesh, Aditya and Goh, Gabriel and Agarwal, Sandhini and Sastry, Girish and Askell, Amanda and Mishkin, Pamela and Clark, Jack and others},
  booktitle={International conference on machine learning},
  pages={8748--8763},
  year={2021},
  organization={PmLR}
}

@inproceedings{li2023blip,
  title={Blip-2: Bootstrapping language-image pre-training with frozen image encoders and large language models},
  author={Li, Junnan and Li, Dongxu and Savarese, Silvio and Hoi, Steven},
  booktitle={International conference on machine learning},
  pages={19730--19742},
  year={2023},
  organization={PMLR}
}

@article{chen2025reasoning,
  title={Reasoning in Space via Grounding in the World},
  author={Chen, Yiming and Qi, Zekun and Zhang, Wenyao and Jin, Xin and Zhang, Li and Liu, Peidong},
  journal={arXiv preprint arXiv:2510.13800},
  year={2025}
}

@article{zheng2025multimodal,
  title={Multimodal spatial reasoning in the large model era: A survey and benchmarks},
  author={Zheng, Xu and Dongfang, Zihao and Jiang, Lutao and Zheng, Boyuan and Guo, Yulong and Zhang, Zhenquan and Albanese, Giuliano and Yang, Runyi and Ma, Mengjiao and Zhang, Zixin and others},
  journal={arXiv preprint arXiv:2510.25760},
  year={2025}
}

@article{bai2025qwen2,
  title={Qwen2. 5-vl technical report},
  author={Bai, Shuai and Chen, Keqin and Liu, Xuejing and Wang, Jialin and Ge, Wenbin and Song, Sibo and Dang, Kai and Wang, Peng and Wang, Shijie and Tang, Jun and others},
  journal={arXiv preprint arXiv:2502.13923},
  year={2025}
}

@article{bai2025qwen3vltechnicalreport,
      title={Qwen3-VL Technical Report}, 
      author={Shuai Bai and Yuxuan Cai and Ruizhe Chen and Keqin Chen and Xionghui Chen and Zesen Cheng and Lianghao Deng and Wei Ding and Chang Gao and Chunjiang Ge and Wenbin Ge and Zhifang Guo and Qidong Huang and Jie Huang and Fei Huang and Binyuan Hui and Shutong Jiang and Zhaohai Li and Mingsheng Li and Mei Li and Kaixin Li and Zicheng Lin and Junyang Lin and Xuejing Liu and Jiawei Liu and Chenglong Liu and Yang Liu and Dayiheng Liu and Shixuan Liu and Dunjie Lu and Ruilin Luo and Chenxu Lv and Rui Men and Lingchen Meng and Xuancheng Ren and Xingzhang Ren and Sibo Song and Yuchong Sun and Jun Tang and Jianhong Tu and Jianqiang Wan and Peng Wang and Pengfei Wang and Qiuyue Wang and Yuxuan Wang and Tianbao Xie and Yiheng Xu and Haiyang Xu and Jin Xu and Zhibo Yang and Mingkun Yang and Jianxin Yang and An Yang and Bowen Yu and Fei Zhang and Hang Zhang and Xi Zhang and Bo Zheng and Humen Zhong and Jingren Zhou and Fan Zhou and Jing Zhou and Yuanzhi Zhu and Ke Zhu},
      journal={arXiv preprint arXiv:2511.21631},
      year={2025}
}

@article{li2024llava,
  title={Llava-onevision: Easy visual task transfer},
  author={Li, Bo and Zhang, Yuanhan and Guo, Dong and Zhang, Renrui and Li, Feng and Zhang, Hao and Zhang, Kaichen and Zhang, Peiyuan and Li, Yanwei and Liu, Ziwei and others},
  journal={arXiv preprint arXiv:2408.03326},
  year={2024}
}

@article{team2024gemini,
  title={Gemini 1.5: Unlocking multimodal understanding across millions of tokens of context},
  author={Team, Gemini and Georgiev, Petko and Lei, Ving Ian and Burnell, Ryan and Bai, Libin and Gulati, Anmol and Tanzer, Garrett and Vincent, Damien and Pan, Zhufeng and Wang, Shibo and others},
  journal={arXiv preprint arXiv:2403.05530},
  year={2024}
}

@article{hurst2024gpt,
  title={Gpt-4o system card},
  author={Hurst, Aaron and Lerer, Adam and Goucher, Adam P and Perelman, Adam and Ramesh, Aditya and Clark, Aidan and Ostrow, AJ and Welihinda, Akila and Hayes, Alan and Radford, Alec and others},
  journal={arXiv preprint arXiv:2410.21276},
  year={2024}
}

@article{zhu2024llava,
  title={Llava-3d: A simple yet effective pathway to empowering lmms with 3d-awareness},
  author={Zhu, Chenming and Wang, Tai and Zhang, Wenwei and Pang, Jiangmiao and Liu, Xihui},
  journal={arXiv preprint arXiv:2409.18125},
  year={2024}
}

@article{qi2025gpt4scene,
  title={Gpt4scene: Understand 3d scenes from videos with vision-language models},
  author={Qi, Zhangyang and Zhang, Zhixiong and Fang, Ye and Wang, Jiaqi and Zhao, Hengshuang},
  journal={arXiv preprint arXiv:2501.01428},
  year={2025}
}

@inproceedings{wang2024dust3r,
  title={Dust3r: Geometric 3d vision made easy},
  author={Wang, Shuzhe and Leroy, Vincent and Cabon, Yohann and Chidlovskii, Boris and Revaud, Jerome},
  booktitle={Proceedings of the IEEE/CVF Conference on Computer Vision and Pattern Recognition},
  pages={20697--20709},
  year={2024}
}

@inproceedings{wang2025continuous,
  title={Continuous 3d perception model with persistent state},
  author={Wang, Qianqian and Zhang, Yifei and Holynski, Aleksander and Efros, Alexei A and Kanazawa, Angjoo},
  booktitle={Proceedings of the Computer Vision and Pattern Recognition Conference},
  pages={10510--10522},
  year={2025}
}

@inproceedings{wang2025vggt,
  title={Vggt: Visual geometry grounded transformer},
  author={Wang, Jianyuan and Chen, Minghao and Karaev, Nikita and Vedaldi, Andrea and Rupprecht, Christian and Novotny, David},
  booktitle={Proceedings of the Computer Vision and Pattern Recognition Conference},
  pages={5294--5306},
  year={2025}
}

@article{wang2025pi,
  title={$\pi^3$: Permutation-Equivariant Visual Geometry Learning},
  author={Wang, Yifan and Zhou, Jianjun and Zhu, Haoyi and Chang, Wenzheng and Zhou, Yang and Li, Zizun and Chen, Junyi and Pang, Jiangmiao and Shen, Chunhua and He, Tong},
  journal={arXiv preprint arXiv:2507.13347},
  year={2025}
}

@article{keetha2025mapanything,
  title={Mapanything: Universal feed-forward metric 3d reconstruction},
  author={Keetha, Nikhil and M{\"u}ller, Norman and Sch{\"o}nberger, Johannes and Porzi, Lorenzo and Zhang, Yuchen and Fischer, Tobias and Knapitsch, Arno and Zauss, Duncan and Weber, Ethan and Antunes, Nelson and others},
  journal={arXiv preprint arXiv:2509.13414},
  year={2025}
}

@article{lin2025depth,
  title={Depth anything 3: Recovering the visual space from any views},
  author={Lin, Haotong and Chen, Sili and Liew, Junhao and Chen, Donny Y and Li, Zhenyu and Shi, Guang and Feng, Jiashi and Kang, Bingyi},
  journal={arXiv preprint arXiv:2511.10647},
  year={2025}
}

@article{li2025iggt,
  title={IGGT: Instance-Grounded Geometry Transformer for Semantic 3D Reconstruction},
  author={Li, Hao and Zou, Zhengyu and Liu, Fangfu and Zhang, Xuanyang and Hong, Fangzhou and Cao, Yukang and Lan, Yushi and Zhang, Manyuan and Yu, Gang and Zhang, Dingwen and others},
  journal={arXiv preprint arXiv:2510.22706},
  year={2025}
}

@article{koch2025unified,
  title={Unified Semantic Transformer for 3D Scene Understanding},
  author={Koch, Sebastian and Wald, Johanna and Matsuki, Hidenobu and Hermosilla, Pedro and Ropinski, Timo and Tombari, Federico},
  journal={arXiv preprint arXiv:2512.14364},
  year={2025}
}

@article{sheng2025spatialsplat,
  title={SpatialSplat: Efficient Semantic 3D from Sparse Unposed Images},
  author={Sheng, Yu and Deng, Jiajun and Zhang, Xinran and Zhang, Yu and Hua, Bei and Zhang, Yanyong and Ji, Jianmin},
  journal={arXiv preprint arXiv:2505.23044},
  year={2025}
}

@article{shen2025fastvggt,
  title={Fastvggt: Training-free acceleration of visual geometry transformer},
  author={Shen, You and Zhang, Zhipeng and Qu, Yansong and Zheng, Xiawu and Ji, Jiayi and Zhang, Shengchuan and Cao, Liujuan},
  journal={arXiv preprint arXiv:2509.02560},
  year={2025}
}

@article{yuan2026infinitevggt,
  title={InfiniteVGGT: Visual Geometry Grounded Transformer for Endless Streams},
  author={Yuan, Shuai and Yang, Yantai and Yang, Xiaotian and Zhang, Xupeng and Zhao, Zhonghao and Zhang, Lingming and Zhang, Zhipeng},
  journal={arXiv preprint arXiv:2601.02281},
  year={2026}
}

@article{wang20254d,
  title={4D-VGGT: A General Foundation Model with SpatioTemporal Awareness for Dynamic Scene Geometry Estimation},
  author={Wang, Haonan and Zhou, Hanyu and Liu, Haoyue and Yan, Luxin},
  journal={arXiv preprint arXiv:2511.18416},
  year={2025}
}

@inproceedings{schonberger2016structure,
  title={Structure-from-motion revisited},
  author={Schonberger, Johannes L and Frahm, Jan-Michael},
  booktitle={Proceedings of the IEEE conference on computer vision and pattern recognition},
  pages={4104--4113},
  year={2016}
}

@book{hartley2003multiple,
  title={Multiple view geometry in computer vision},
  author={Hartley, Richard and Zisserman, Andrew},
  year={2003},
  publisher={Cambridge university press}
}

@article{wu2025spatial,
  title={Spatial-mllm: Boosting mllm capabilities in visual-based spatial intelligence},
  author={Wu, Diankun and Liu, Fangfu and Hung, Yi-Hsin and Duan, Yueqi},
  journal={arXiv preprint arXiv:2505.23747},
  year={2025}
}

@article{feng2025video,
  title={Video-r1: Reinforcing video reasoning in mllms},
  author={Feng, Kaituo and Gong, Kaixiong and Li, Bohao and Guo, Zonghao and Wang, Yibing and Peng, Tianshuo and Wu, Junfei and Zhang, Xiaoying and Wang, Benyou and Yue, Xiangyu},
  journal={arXiv preprint arXiv:2503.21776},
  year={2025}
}

@inproceedings{yang2025thinking,
  title={Thinking in space: How multimodal large language models see, remember, and recall spaces},
  author={Yang, Jihan and Yang, Shusheng and Gupta, Anjali W and Han, Rilyn and Fei-Fei, Li and Xie, Saining},
  booktitle={Proceedings of the Computer Vision and Pattern Recognition Conference},
  pages={10632--10643},
  year={2025}
}

@article{yang2025visual,
  title={Visual spatial tuning},
  author={Yang, Rui and Zhu, Ziyu and Li, Yanwei and Huang, Jingjia and Yan, Shen and Zhou, Siyuan and Liu, Zhe and Li, Xiangtai and Li, Shuangye and Wang, Wenqian and others},
  journal={arXiv preprint arXiv:2511.05491},
  year={2025}
}

@inproceedings{SpatialLM,
  title     = {SpatialLM: Training Large Language Models for Structured Indoor Modeling},
  author    = {Mao, Yongsen and Zhong, Junhao and Fang, Chuan and Zheng, Jia and Tang, Rui and Zhu, Hao and Tan, Ping and Zhou, Zihan},
  booktitle = {Advances in Neural Information Processing Systems},
  year      = {2025}
}

@inproceedings{zheng2025video,
  title={Video-3d llm: Learning position-aware video representation for 3d scene understanding},
  author={Zheng, Duo and Huang, Shijia and Wang, Liwei},
  booktitle={Proceedings of the Computer Vision and Pattern Recognition Conference},
  pages={8995--9006},
  year={2025}
}

@article{zheng2025learning,
  title={Learning from Videos for 3D World: Enhancing MLLMs with 3D Vision Geometry Priors},
  author={Zheng, Duo and Huang, Shijia and Li, Yanyang and Wang, Liwei},
  journal={arXiv preprint arXiv:2505.24625},
  year={2025}
}

@article{fan2025vlm,
  title={VLM-3R: Vision-Language Models Augmented with Instruction-Aligned 3D Reconstruction},
  author={Fan, Zhiwen and Zhang, Jian and Li, Renjie and Zhang, Junge and Chen, Runjin and Hu, Hezhen and Wang, Kevin and Qu, Huaizhi and Wang, Dilin and Yan, Zhicheng and others},
  journal={arXiv preprint arXiv:2505.20279},
  year={2025}
}

@article{cai2025depthlm,
  title={Depthlm: Metric depth from vision language models},
  author={Cai, Zhipeng and Yeh, Ching-Feng and Xu, Hu and Liu, Zhuang and Meyer, Gregory and Lei, Xinjie and Zhao, Changsheng and Li, Shang-Wen and Chandra, Vikas and Shi, Yangyang},
  journal={arXiv preprint arXiv:2509.25413},
  year={2025}
}

@article{zhang2025flatland,
  title={From flatland to space: Teaching vision-language models to perceive and reason in 3d},
  author={Zhang, Jiahui and Chen, Yurui and Zhou, Yanpeng and Xu, Yueming and Huang, Ze and Mei, Jilin and Chen, Junhui and Yuan, Yu-Jie and Cai, Xinyue and Huang, Guowei and others},
  journal={arXiv preprint arXiv:2503.22976},
  year={2025}
}

@inproceedings{dai2017scannet,
  title={Scannet: Richly-annotated 3d reconstructions of indoor scenes},
  author={Dai, Angela and Chang, Angel X and Savva, Manolis and Halber, Maciej and Funkhouser, Thomas and Nie{\ss}ner, Matthias},
  booktitle={Proceedings of the IEEE conference on computer vision and pattern recognition},
  pages={5828--5839},
  year={2017}
}

@inproceedings{wang2024embodiedscan,
  title={Embodiedscan: A holistic multi-modal 3d perception suite towards embodied ai},
  author={Wang, Tai and Mao, Xiaohan and Zhu, Chenming and Xu, Runsen and Lyu, Ruiyuan and Li, Peisen and Chen, Xiao and Zhang, Wenwei and Chen, Kai and Xue, Tianfan and others},
  booktitle={Proceedings of the IEEE/CVF Conference on Computer Vision and Pattern Recognition},
  pages={19757--19767},
  year={2024}
}

@inproceedings{scanrefer,
 author = {Dave Zhenyu Chen and
Angel X. Chang and
Matthias Nie{\ss}ner},
 booktitle = {ECCV},
 title = {ScanRefer: 3D Object Localization in {RGB-D} Scans Using Natural Language},
 year = {2020}
}

@inproceedings{chen2021scan2cap,
  title={Scan2cap: Context-aware dense captioning in rgb-d scans},
  author={Chen, Zhenyu and Gholami, Ali and Nie{\ss}ner, Matthias and Chang, Angel X},
  booktitle={Proceedings of the IEEE/CVF conference on computer vision and pattern recognition},
  pages={3193--3203},
  year={2021}
}

@article{huang2023embodied,
  title={An embodied generalist agent in 3d world},
  author={Huang, Jiangyong and Yong, Silong and Ma, Xiaojian and Linghu, Xiongkun and Li, Puhao and Wang, Yan and Li, Qing and Zhu, Song-Chun and Jia, Baoxiong and Huang, Siyuan},
  journal={arXiv preprint arXiv:2311.12871},
  year={2023}
}

@article{huang2024chat,
  title={Chat-scene: Bridging 3d scene and large language models with object identifiers},
  author={Huang, Haifeng and Chen, Yilun and Wang, Zehan and Huang, Rongjie and Xu, Runsen and Wang, Tai and Liu, Luping and Cheng, Xize and Zhao, Yang and Pang, Jiangmiao and others},
  journal={Advances in Neural Information Processing Systems},
  volume={37},
  pages={113991--114017},
  year={2024}
}

@article{chen2024grounded,
  title={Grounded 3d-llm with referent tokens},
  author={Chen, Yilun and Yang, Shuai and Huang, Haifeng and Wang, Tai and Xu, Runsen and Lyu, Ruiyuan and Lin, Dahua and Pang, Jiangmiao},
  journal={arXiv preprint arXiv:2405.10370},
  year={2024}
}

@inproceedings{chen2024ll3da,
  title={Ll3da: Visual interactive instruction tuning for omni-3d understanding reasoning and planning},
  author={Chen, Sijin and Chen, Xin and Zhang, Chi and Li, Mingsheng and Yu, Gang and Fei, Hao and Zhu, Hongyuan and Fan, Jiayuan and Chen, Tao},
  booktitle={Proceedings of the IEEE/CVF conference on computer vision and pattern recognition},
  pages={26428--26438},
  year={2024}
}

@article{hong20233d,
  title={3d-llm: Injecting the 3d world into large language models},
  author={Hong, Yining and Zhen, Haoyu and Chen, Peihao and Zheng, Shuhong and Du, Yilun and Chen, Zhenfang and Gan, Chuang},
  journal={Advances in Neural Information Processing Systems},
  volume={36},
  pages={20482--20494},
  year={2023}
}

@article{meng2024deepstack,
  title={Deepstack: Deeply stacking visual tokens is surprisingly simple and effective for lmms},
  author={Meng, Lingchen and Yang, Jianwei and Tian, Rui and Dai, Xiyang and Wu, Zuxuan and Gao, Jianfeng and Jiang, Yu-Gang},
  journal={Advances in Neural Information Processing Systems},
  volume={37},
  pages={23464--23487},
  year={2024}
}

@inproceedings{achlioptas2020referit3d,
  title={Referit3d: Neural listeners for fine-grained 3d object identification in real-world scenes},
  author={Achlioptas, Panos and Abdelreheem, Ahmed and Xia, Fei and Elhoseiny, Mohamed and Guibas, Leonidas},
  booktitle={European conference on computer vision},
  pages={422--440},
  year={2020},
  organization={Springer}
}

@inproceedings{chen2022d3net,
  title={D 3 net: A unified speaker-listener architecture for 3d dense captioning and visual grounding},
  author={Chen, Dave Zhenyu and Wu, Qirui and Nie{\ss}ner, Matthias and Chang, Angel X},
  booktitle={European Conference on Computer Vision},
  pages={487--505},
  year={2022},
  organization={Springer}
}

@inproceedings{chen2023unit3d,
  title={Unit3d: A unified transformer for 3d dense captioning and visual grounding},
  author={Chen, Zhenyu and Hu, Ronghang and Chen, Xinlei and Nie{\ss}ner, Matthias and Chang, Angel X},
  booktitle={Proceedings of the IEEE/CVF international conference on computer vision},
  pages={18109--18119},
  year={2023}
}

@article{cao2026vggtdet,
  title={VGGT-Det: Mining VGGT Internal Priors for Sensor-Geometry-Free Multi-View Indoor 3D Object Detection},
  author={Cao, Yang and Wu, Feize and Chen, Dave Zhenyu and Zhong, Yingji and Hong, Lanqing and Xu, Dan},
  journal={arXiv preprint arXiv:2603.00912},
  year={2026}
}

@article{cao2023coda,
  title={Coda: Collaborative novel box discovery and cross-modal alignment for open-vocabulary 3d object detection},
  author={Cao, Yang and Yihan, Zeng and Xu, Hang and Xu, Dan},
  journal={Advances in Neural Information Processing Systems},
  volume={36},
  pages={71862--71873},
  year={2023}
}

@article{cao2025codav2,
  title={Collaborative novel object discovery and box-guided cross-modal alignment for open-vocabulary 3d object detection},
  author={Cao, Yang and Zeng, Yihan and Xu, Hang and Xu, Dan},
  journal={IEEE Transactions on Pattern Analysis and Machine Intelligence},
  year={2025},
  publisher={IEEE}
}

@article{cao20243dgsdet,
  title={3dgs-det: Empower 3d gaussian splatting with boundary guidance and box-focused sampling for 3d object detection},
  author={Cao, Yang and Jv, Yuanliang and Xu, Dan},
  journal={arXiv preprint arXiv:2410.01647},
  year={2024}
}

@inproceedings{qi2019votenet,
  title={Deep hough voting for 3d object detection in point clouds},
  author={Qi, Charles R and Litany, Or and He, Kaiming and Guibas, Leonidas J},
  booktitle={proceedings of the IEEE/CVF International Conference on Computer Vision},
  pages={9277--9286},
  year={2019}
}

@inproceedings{qi2020imvotenet,
  title={Imvotenet: Boosting 3d object detection in point clouds with image votes},
  author={Qi, Charles R and Chen, Xinlei and Litany, Or and Guibas, Leonidas J},
  booktitle={Proceedings of the IEEE/CVF conference on computer vision and pattern recognition},
  pages={4404--4413},
  year={2020}
}

@article{ma2024whenllms,
  title={When llms step into the 3d world: A survey and meta-analysis of 3d tasks via multi-modal large language models},
  author={Ma, Xianzheng and Smart, Brandon and Bhalgat, Yash and Chen, Shuai and Li, Xinghui and Ding, Jian and Gu, Jindong and Chen, Dave Zhenyu and Peng, Songyou and Bian, Jia-Wang and others},
  journal={arXiv preprint arXiv:2405.10255},
  year={2024}
}

@article{xu2025uniugg,
  title={Uniugg: Unified 3d understanding and generation via geometric-semantic encoding},
  author={Xu, Yueming and Zhang, Jiahui and Huang, Ze and Chen, Yurui and Zhou, Yanpeng and Chen, Zhenyu and Yuan, Yu-Jie and Xia, Pengxiang and Huang, Guowei and Cai, Xinyue and others},
  journal={arXiv preprint arXiv:2508.11952},
  year={2025}
}

@article{dwedari2023generating,
  title={Generating context-aware natural answers for questions in 3D scenes},
  author={Dwedari, Mohammed Munzer and Niessner, Matthias and Chen, Dave Zhenyu},
  journal={arXiv preprint arXiv:2310.19516},
  year={2023}
}

@article{gao2026map2thought,
  title={Map2Thought: Explicit 3D Spatial Reasoning via Metric Cognitive Maps},
  author={Gao, Xiangjun and Zhang, Zhensong and Chen, Dave Zhenyu and Xu, Songcen and Quan, Long and P{\'e}rez-Pellitero, Eduardo and Jang, Youngkyoon},
  journal={arXiv preprint arXiv:2601.11442},
  year={2026}
}

@article{nuscenes2019,
  title={nuScenes: A multimodal dataset for autonomous driving},
  author={Holger Caesar and Varun Bankiti and Alex H. Lang and Sourabh Vora and 
          Venice Erin Liong and Qiang Xu and Anush Krishnan and Yu Pan and 
          Giancarlo Baldan and Oscar Beijbom},
  journal={arXiv preprint arXiv:1903.11027},
  year={2019}
}

@inproceedings{li2026causal,
  title={Causal tracing of object representations in large vision language models: Mechanistic interpretability and hallucination mitigation},
  author={Li, Qiming and Ye, Zekai and Feng, Xiaocheng and Zhong, Weihong and Ma, Weitao and Feng, Xiachong},
  booktitle={Proceedings of the AAAI Conference on Artificial Intelligence},
  pages={31645--31653},
  year={2026}
}

\clearpage
\setcounter{section}{5}
\setcounter{table}{8}
\setcounter{figure}{9}
\maketitlesupplementary

This supplementary material provides additional details to complement the main paper.
We first describe the training prompt design in Sec.~\ref{sec:dataset_prompt}, covering both sparse geometry--semantics joint pre-training and downstream task fine-tuning.
We then present extended ablation studies in Sec.~\ref{sec:analysis} to further analyze the contributions of each design component.
Additional qualitative visualizations are provided in Sec.~\ref{sec:qualitative} to illustrate the learned geometric perception ability.
Finally, we discuss the limitations of the current framework and possible future directions in Sec.~\ref{sec:limitation}.

\section{Training Prompt Design}
\label{sec:dataset_prompt}
In this section, we provide the complete prompts used in both sparse geometry–semantics joint pre-training and object-level 3D perception fine-tuning.

\subsection{Sparse Geometry–Semantics Joint Pre-training}
During sparse joint pre-training, a red cross is used as a visual prompt to mark a target pixel in one frame of the input image sequence. The model is required to predict both the semantic label and the 3D coordinate of the pixel under the unified first-frame coordinate system, as shown in Tab.~\ref{tab:prompt_sparse}.

\begin{table}[htbp]
\centering
\caption{\textbf{The prompt for sparse geometry--semantics joint pre-training.} Given multi-frame images and a queried pixel marked by a red cross, the model predicts the semantic label and the 3D coordinate of the point in the first-frame coordinate system.}
\begin{tcolorbox}[
  enhanced,
  width=\linewidth,
  colback=gray!10,
  colframe=black!70,
  boxrule=0.8pt,
  arc=3.5mm,
  left=2.5mm,right=2.5mm,top=2mm,bottom=2mm,
  boxsep=1.2mm,
  before upper=\raggedright,
  fontupper=\small,
]

\tasktitle{Sparse Geometry--Semantics Joint Pre-training}
\smallskip

\Human\ \imgtok\markedimgtok\imgtok\imgtok\imgdots

Given the images, find the point covered by the red cross.

Output a JSON dictionary with the point's semantic label in \texttt{"label"}
and the point's 3D coordinate in \texttt{"pointmap"} in the camera coordinate
system of the first frame.\par

\GPT\par
\begin{lstlisting}[style=jsonstyle]
(*@\fencejson@*)
{"label": "monitor", "pointmap": [-0.32, -0.54, 1.69]}
(*@\fence@*)
\end{lstlisting}

\end{tcolorbox}

\label{tab:prompt_sparse}
\end{table}

\subsection{3D Visual Grounding}
The two-stage grounding used for 3D visual grounding is illustrated in Tab.~\ref{tab:prompt_grounding}.
In the first stage, the model selects the most informative frame that clearly observes the target object.
In the second stage, a set of frames is sampled around the selected anchor frame, while the anchor frame is placed as the first frame to define the coordinate system.
This design enforces a unified coordinate system across different tasks and strengthens spatial consistency during grounding.
The quantitative impact of this design is analyzed in Sec.~\ref{supp:Two-Stage Grounding}.

\begin{table}[htbp]
\centering
\caption{\textbf{Two-stage prompt for 3D visual grounding.}
The first stage selects the most informative frame for grounding,
and the second stage predicts the 3D bounding box under the
coordinate system of the selected frame.}

\begin{tcolorbox}[
  enhanced,
  width=\linewidth,
  colback=gray!10,
  colframe=black!70,
  boxrule=0.8pt,
  arc=3.5mm,
  left=2.5mm,right=2.5mm,top=2mm,bottom=2mm,
  boxsep=1.2mm,
  before upper=\raggedright,
  fontupper=\small,
]

\tasktitle{Stage 1: Anchor Frame Selection}
\smallskip

\Human\ Frame-0: \imgtok Frame-1: \imgtok Frame-2: \imgtok \imgdots

Localize the first clear frame in the video showing the object described in the text.

Text: A white cabinet in the corner of the room. In the direction from the door and from the inside, it will be on the left.

Output a JSON dictionary with the frame index in \texttt{"frame"}.

\GPT

\textasciigrave\textasciigrave\textasciigrave json \\
\texttt{\{"frame": 2\}} \\
\textasciigrave\textasciigrave\textasciigrave

\tcbline

\tasktitle{Stage 2: Anchor-Conditioned Localization}
\smallskip

\Human\ \starimgtok \imgtok \imgtok \imgtok \imgdots

Localize the object described in the text.

Text: A white cabinet in the corner of the room. In the direction from the door and from the inside, it will be on the left.

Output a JSON dictionary with the matched object's 3D bounding box in \texttt{"bbox\_3d"}
in the camera coordinate system of the first frame.

The 3D bounding box format should be
\texttt{[x\_center, y\_center, z\_center, x\_size, y\_size, z\_size, yaw, pitch, roll]}.

\GPT

\textasciigrave\textasciigrave\textasciigrave json \\
\texttt{\{"bbox\_3d": [-0.39, -0.19, 2.7, 0.6, 1.22, 1.94, 0.79, 1.32, -3.07]\}} \\
\textasciigrave\textasciigrave\textasciigrave

\end{tcolorbox}

\label{tab:prompt_grounding}

\end{table}

\subsection{3D Dense Captioning}

The prompt for 3D dense captioning is shown in Tab.~\ref{tab:prompt_caption}.
Given multi-frame images and a queried 3D coordinate in the first-frame coordinate system, the model generates a natural language description of the corresponding object.
This formulation follows the same unified coordinate definition as other tasks.

\begin{table}[htbp]
\centering
\vspace{-3mm}
\caption{\textbf{The prompt for 3D dense captioning.}
Given multi-frame images and a queried 3D point in the first-frame coordinate system,
the model generates a natural language description of the corresponding object.}
\vspace{-2mm}
\begin{tcolorbox}[
  enhanced,
  width=\linewidth,
  colback=gray!10,
  colframe=black!70,
  boxrule=0.8pt,
  arc=3.5mm,
  left=2.5mm,right=2.5mm,top=2mm,bottom=2mm,
  boxsep=1.2mm,
  before upper=\raggedright,
  fontupper=\small,
]

\tasktitle{3D Dense Captioning}
\smallskip

\Human\ \imgtok\imgtok\imgtok\imgtok\imgdots

Carefully watch the video and describe the object located at \texttt{[3.45, 0.9, -1.37]} in detail.

\GPT

A white cabinet in the corner of the room. In the direction from the door and from the inside, it will be on the left. 

\end{tcolorbox}

\label{tab:prompt_caption}
\end{table}

\vspace{-6mm}

\subsection{3D Video Object Detection}
The task prompt used for 3D video object detection is shown in Tab.~\ref{tab:prompt_detection}.
Given multi-frame images, the model is required to predict all objects and their 3D bounding boxes under the first-frame coordinate system.
This formulation follows the same unified coordinate definition as other downstream tasks.

\begin{table}[htbp]
\centering
\vspace{-3mm}
\caption{\textbf{Task prompt for 3D video object detection.} Given multi-frame images, the model predicts all object categories and their 3D bounding boxes in the first-frame coordinate system.}
\vspace{-2mm}
\begin{tcolorbox}[
  enhanced,
  width=\linewidth,
  colback=gray!10,
  colframe=black!70,
  boxrule=0.8pt,
  arc=3.5mm,
  left=2.5mm,right=2.5mm,top=2mm,bottom=2mm,
  boxsep=1.2mm,
  before upper=\raggedright,
  fontupper=\small,
]

\tasktitle{3D Video Object Detection}
\smallskip

\Human\ \imgtok\imgtok\imgtok\imgtok\imgdots

Detect the 3D bounding boxes in the camera coordinate system of the first frame.

Output a JSON list where each entry contains the object name in \texttt{"label"} and its 3D bounding box in \texttt{"bbox\_3d"}.

The 3D bounding box format should be
\texttt{[x\_center, y\_center, z\_center, x\_size, y\_size, z\_size, yaw, pitch, roll]}.

\GPT \\

\textasciigrave\textasciigrave\textasciigrave json \\
\texttt{[} \\
$\qquad$ \texttt{\{"label": "table", "bbox\_3d": [0.32, 0.6, 1.05, 0.86, 1.7, 0.87, -1.34, 1.08, -2.84]\},} \\
$\qquad$ \texttt{\dots} \\
\texttt{]} \\
\textasciigrave\textasciigrave\textasciigrave

\end{tcolorbox}

\label{tab:prompt_detection}
\end{table}

\section{Additional Ablation Study}
\label{sec:analysis}

\subsection{Detailed Ablations on Core Contributions}
\label{supp:Core Contribution}
We evaluate the effectiveness of the proposed sparse geometry--semantics joint pre-training (A) and the multi-level gated fusion module (B) across different backbones and tasks.
The complete results are summarized in Tab.~\ref{tab:ablation_detection}.

Our ablation experiments are conducted on two MLLM frameworks, VG-LLM-4B~\cite{zheng2025learning} and ours, and evaluated on three representative tasks, including 3D visual grounding (ScanRefer~\cite{scanrefer}), 3D dense captioning (Scan2Cap~\cite{chen2021scan2cap}), and 3D video object detection.
For GAP-MLLM, the baseline with neither sparse joint pre-training nor multi-level fusion follows VG-LLM~\cite{zheng2025learning} and fuses last-layer geometric tokens with final visual tokens by element-wise addition.

Overall, both components consistently improve performance across tasks and models.
Sparse joint pre-training yields clear gains on all three tasks, showing that explicitly activating geometric perception benefits downstream 3D perception.
Multi-level gated fusion further improves the interaction between geometric and visual features, bringing additional gains.

Combining both gives the best results across all settings.
For example, under GAP-MLLM-3B, the full model improves ScanRefer accuracy from 47.9/22.6 to 53.0/26.0 and raises the F1 score of 3D video detection from 44.7 to 50.6.
These results demonstrate that geometry-aligned pre-training and progressive feature fusion provide complementary benefits for improving 3D perception in MLLMs.

\begin{table*}[htbp]
\scriptsize
\centering
\caption{
\textbf{Ablation on sparse joint pre-training (A) and multi-level gated fusion (B) across multiple tasks.}
}
\setlength{\tabcolsep}{3mm}
\renewcommand{\arraystretch}{1.1}

\begin{tabular}{cc|cc|cc|ccc}
\toprule
\multirow{2}{*}{A} & \multirow{2}{*}{B}
& \multicolumn{2}{c|}{\textbf{ScanRefer}}
& \multicolumn{2}{c|}{\textbf{Scan2Cap}}
& \multicolumn{3}{c}{\textbf{3D Video Detection}} \\

\cmidrule(lr){3-4} \cmidrule(lr){5-6} \cmidrule(lr){7-9}

& 
& Acc@0.25 & Acc@0.5
& C@0.5 & B-4@0.5
& P$_{25}$ & R$_{25}$ & F1$_{25}$ \\

\midrule
\multicolumn{9}{c}{VG-LLM-4B (Qwen2.5-VL + VGGT)} \\
\midrule

 & 
& 45.4 & 19.3 
& 77.0 & 40.6
& 41.2 & 35.9 & 37.9 \\

\checkmark & 
& 48.2 & 22.6
& 79.7 & 40.8
& 44.1 & 39.6 & 41.4 \\

 & \checkmark
& 46.8 & 20.5
& 77.5 & 40.6
& 43.3 & 36.9 & 39.4 \\

\checkmark & \checkmark
& \textbf{49.7} & \textbf{23.2}
& \textbf{80.6} & \textbf{41.1}
& \textbf{47.0} & \textbf{41.2} & \textbf{43.5} \\

\midrule
\multicolumn{9}{c}{GAP-MLLM-3B (Qwen3-VL + VGGT)} \\
\midrule

 & 
& 47.9 & 22.6
& 79.9 & 41.2
& 48.8 & 42.1 & 44.7 \\

\checkmark & 
& 51.7 & 25.1
& 82.5 & 41.9
& 52.7 & 45.9 & 48.7 \\

 & \checkmark
& 51.1 & 23.9
& 82.7 & 41.8
& 51.8 & 44.6 & 47.5 \\

\checkmark & \checkmark
& \textbf{53.0} & \textbf{26.0}
& \textbf{84.7} & \textbf{42.1}
& \textbf{54.2} & \textbf{48.1} & \textbf{50.6} \\

\bottomrule
\end{tabular}

\label{tab:ablation_detection}
\end{table*}

\vspace{-8mm}

\subsection{Effect of Two-Stage Grounding}
\label{supp:Two-Stage Grounding}
Tab.~\ref{tab:twostage_effect} shows the effect of the proposed two-stage grounding framework.
Compared with the original VG-LLM formulation, introducing two-stage grounding significantly improves grounding performance (e.g., Acc@0.25 from 37.5 to 45.4 on ScanRefer).

This improvement mainly comes from decoupling frame selection and localization, as well as adopting a unified coordinate system across tasks.
We also observe small but consistent improvements in captioning and detection, highlighting the benefit of unified spatial representation in multi-task 3D perception.

All experiments of GAP-MLLM reported in the main paper, as well as the ablation results in Tab.~\ref{tab:ablation_detection}, are conducted under this two-stage grounding formulation.

\begin{table}[t]
\centering
\scriptsize
\setlength{\tabcolsep}{4pt}
\renewcommand{\arraystretch}{1.15}
\caption{
\textbf{Effect of Two-Stage Grounding under the unified coordinate system.}
Introducing Two-Stage Grounding consistently improves performance
across multiple 3D perception tasks under the unified coordinate system.
}
\begin{tabular}{l|cc|cc|ccc}
\toprule

\multirow{2}{*}{\textbf{Method}} 
& \multicolumn{2}{c|}{\textbf{ScanRefer}}
& \multicolumn{2}{c|}{\textbf{Scan2Cap}}
& \multicolumn{3}{c}{\textbf{3D Video Detection}} \\

\cmidrule(lr){2-3}
\cmidrule(lr){4-5}
\cmidrule(lr){6-8}

& Acc@0.25 & Acc@0.5
& C@0.5 & B-4@0.5
& P$_{25}$ & R$_{25}$ & F1$_{25}$ \\

\midrule

VG-LLM
& 37.5 & 11.1
& 76.9 & 40.5
& 40.3 & 34.9 & 36.9 \\

VG-LLM + StageGrounding
& \textbf{45.4} & \textbf{19.3}
& \textbf{77.0} & \textbf{40.6}
& \textbf{41.2} & \textbf{35.9} & \textbf{37.9} \\

\bottomrule
\end{tabular}


\label{tab:twostage_effect}
\end{table}

\subsection{Robustness to Degraded First Frame}

We evaluate model robustness under degraded initial observations, where the first frame is artificially corrupted while subsequent frames remain unchanged. We consider two types of perturbations: Gaussian blur and partial occlusion. Blur is controlled by the kernel radius, while occlusion is applied by masking a rectangular region with varying location and size.

\begin{table}[htbp]
\centering
\small
\setlength{\tabcolsep}{4pt}
\renewcommand{\arraystretch}{1.15}

\caption{\textbf{Robustness under first-frame degradation.} Blur severity is controlled by Gaussian kernel standard deviation $\sigma$, and occlusion severity is defined as the fraction of masked image area.}
\label{tab:first_frame_robustness}

\begin{tabular}{lccc}
\toprule
\textbf{Setting} & Precision & Recall & F1 \\
\midrule
GAP-MLLM & 54.2 & 48.1 & 50.6 \\
Blur ($\sigma=8$) & 52.1 & 44.4 & 47.5 \\
Blur ($\sigma=12$) & 50.7 & 42.9 & 46.0 \\
Occlusion (30\%) & 53.8 & 47.6 & 50.1 \\
Occlusion (50\%) & 51.9 & 46.0 & 48.4 \\
\bottomrule
\end{tabular}
\end{table}

\begin{figure}[htbp]
\centering
\includegraphics[width=\linewidth]{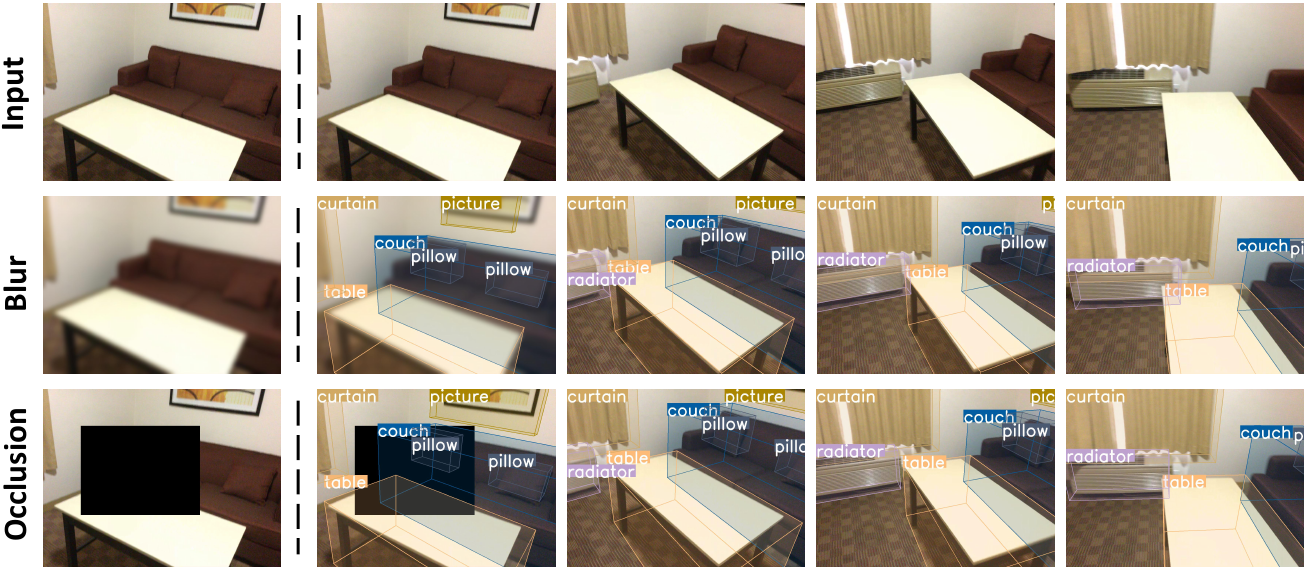}
\caption{\textbf{Qualitative results under first-frame degradation.} Even when the first frame is severely corrupted (e.g., blur or occlusion), GAP-MLLM can still recover coherent object-level 3D structure by leveraging temporal information from subsequent frames, demonstrating strong robustness to unreliable initial observations.}
\vspace{-3mm}
\label{fig:first_frame_robustness}
\end{figure}

As shown in Table~\ref{tab:first_frame_robustness}, GAP-MLLM maintains stable performance under all degradation settings. Even with severe blur or occlusion, performance drops remain limited, indicating robustness to unreliable initial observations and effective multi-frame geometric aggregation. Figure~\ref{fig:first_frame_robustness} shows that the model can still recover coherent object-level 3D structures even when the first frame is heavily degraded.

\subsection{Impact of Patch Merging on Small Objects}

Patch merging reduces token cost by compressing spatial tokens, which may potentially affect fine-grained object representation, especially for small objects. To evaluate this effect, we analyze performance on eight representative small-object categories under token compression.

We report both per-category performance and overall averaged metrics. The evaluation includes eight small-object classes. This allows us to assess whether token compression introduces bias toward large or salient objects.

As shown in Table~\ref{tab:small_object_results}, GAP-MLLM significantly outperforms VG-LLM~\cite{zheng2025learning} across most categories and achieves consistent gains in average precision, recall, and F1 score. This demonstrates that multi-level geometric fusion effectively preserves small-object information even after patch merging.

\begin{table}[htbp]
\centering
\scriptsize
\renewcommand{\arraystretch}{1.05}
\setlength{\tabcolsep}{3.5pt}
\vspace{-3mm}
\caption{\textbf{Performance on small-object categories under patch merging.} The table reports per-category F1 scores and overall averaged metrics over eight categories. TP = Toilet paper, TB = Tissue box, HD = Hair dryer, Disp = Dispenser.}
\label{tab:small_object_results}

\begin{tabular}{l|cccccccc|ccc}
\toprule

\multirow{2}{*}{\textbf{Method}}
& \multicolumn{8}{c|}{\textbf{Small-object F1} (per category)}
& \multicolumn{3}{c}{\textbf{Average metrics}} \\
\cmidrule(lr){2-9} \cmidrule(lr){10-12}

& Soap & TP & TB & Pillow & Case & Plant & HD & Disp
& P$_{25}$ & R$_{25}$ & F1$_{25}$ \\
\midrule

VG-LLM
& 11.3 & 18.0 & 2.4 & 14.8 & 34.9 & 15.3 & 0.0 & 31.4
& 18.0 & 15.2 & 16.0 \\

GAP-MLLM
& 44.4 & 32.4 & 15.2 & 28.4 & 53.2 & 19.6 & 12.5 & 43.6
& 36.4 & 29.7 & 31.2 \\

\bottomrule
\end{tabular}
\end{table}
\vspace{-3mm}

\begin{figure}[t]
\centering
\includegraphics[width=\linewidth]{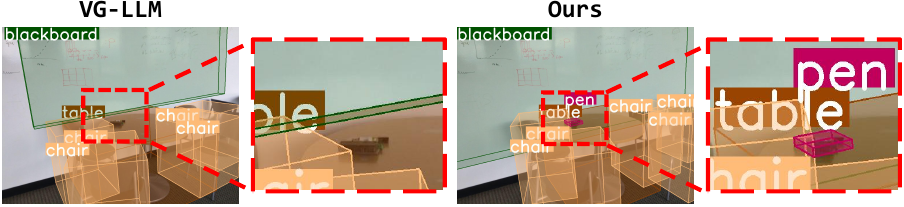}

\caption{\textbf{Effect of patch merging on small-object perception.}
Compared with VG-LLM, GAP-MLLM better preserves small-object details under token compression.
This indicates that multi-level geometric fusion effectively mitigates information loss caused by patch merging.}
\label{fig:small_object_qualitative}
\end{figure}

Figure~\ref{fig:small_object_qualitative} further provides qualitative comparisons. Even under token compression, GAP-MLLM preserves fine-grained object localization and avoids missing small or low-visibility objects, whereas VG-LLM~\cite{zheng2025learning} often fails to detect or precisely localize them.

\subsection{Dynamic Scenes and Domain Shifts}

Our default setting assumes that input frames describe the same scene state, and thus does not explicitly model object motion. To extend GAP-MLLM to dynamic scenes, we adopt a sliding-window inference strategy. Given a video sequence, each local window is processed independently, and the predicted 3D boxes are represented in the coordinate system of the first frame of that window.

\begin{figure}[htbp]
\centering
\includegraphics[width=\linewidth]{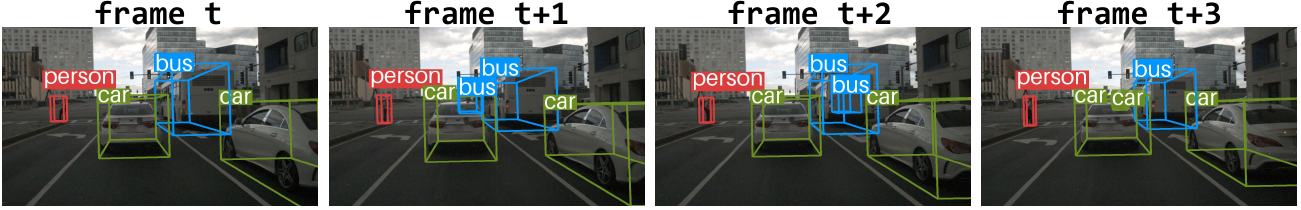}
\caption{\textbf{Dynamic-scene extension with sliding-window inference.}
Each local window predicts 3D boxes in the coordinate system of its first frame. GAP-MLLM produces plausible object-level predictions on NuScenes, showing preliminary generalization to dynamic outdoor scenes.}
\label{fig:dynamic_scene}
\end{figure}

As shown in Figure~\ref{fig:dynamic_scene}, GAP-MLLM produces plausible 3D detection boxes on NuScenes~\cite{nuscenes2019} outdoor driving scenes, including dynamic objects such as cars, pedestrians, and buses. This suggests that the geometric prior inherited from VGGT provides a certain degree of cross-domain generalization. Nevertheless, since temporal association and object motion are not explicitly modeled, dynamic 3D perception remains an important direction for future work.

\vspace{-3mm}
\section{Additional Qualitative Results}
\label{sec:qualitative}

In this section, we provide additional qualitative results to examine the capabilities of our method.
We visualize three representative aspects of GAP-MLLM: metric reconstruction, 3D visual grounding, and 3D video object detection.
These examples illustrate the geometric perception ability learned by our geometry-aligned pre-training and its transferability across multiple downstream tasks.

\begin{figure*}[t]
\centering
\includegraphics[width=\linewidth]{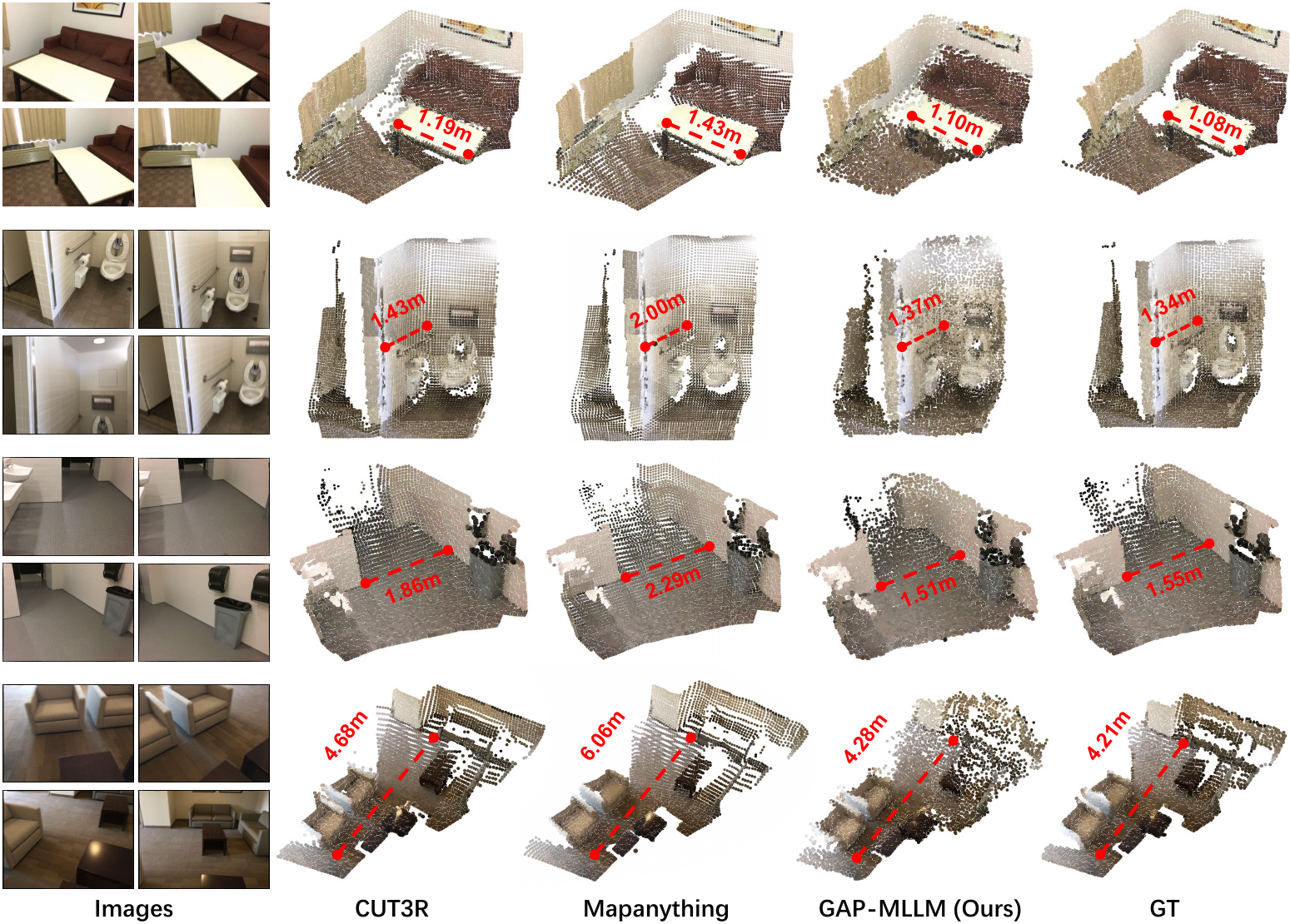}
\caption{
\textbf{Qualitative comparison on metric reconstruction.}
Given multi-view RGB inputs (left), different models reconstruct 3D metric point clouds, where the red line indicates the measured distance.
Our method produces metric distances closest to the ground-truth scale while preserving coherent geometry, demonstrating that geometry-aligned pre-training enables MLLMs to generate metrically pointmap.
}
\label{fig:metric_reconstruction}
\end{figure*}

\subsection{Metric Reconstruction}
\label{supp:metric_reconstruction}

Metric reconstruction provides a direct probe of whether the model has acquired metric-aware 3D perception after geometry-aligned pre-training.
Our model predicts 3D geometry through pixel-level queries, where the MLLM directly outputs the 3D coordinate for each queried pixel, forming a dense pointmap under the coordinate system of the first frame.

An important property of our reconstruction is that the predicted geometry is metric-consistent.
The generated point clouds preserve real-world scale and are directly aligned with the first-frame coordinate system.
This property is particularly important for downstream 3D perception tasks, as it provides a unified spatial representation for grounding, detection, and captioning.

Fig.~\ref{fig:metric_reconstruction} shows qualitative comparisons with existing feed-forward metric reconstruction models~\cite{wang2025continuous, keetha2025mapanything}.
For several representative scenes, we measure the distance between two points in the reconstructed point cloud and compare it with the ground truth.
Our results produce metric distances that are consistently closer to the ground-truth scale while maintaining coherent scene geometry, demonstrating that sparse geometry--semantics joint pre-training enables MLLMs to recover accurate metric 3D structure from RGB inputs.

\subsection{3D Visual Grounding}

\begin{figure*}[t]
\centering
\includegraphics[width=\linewidth]{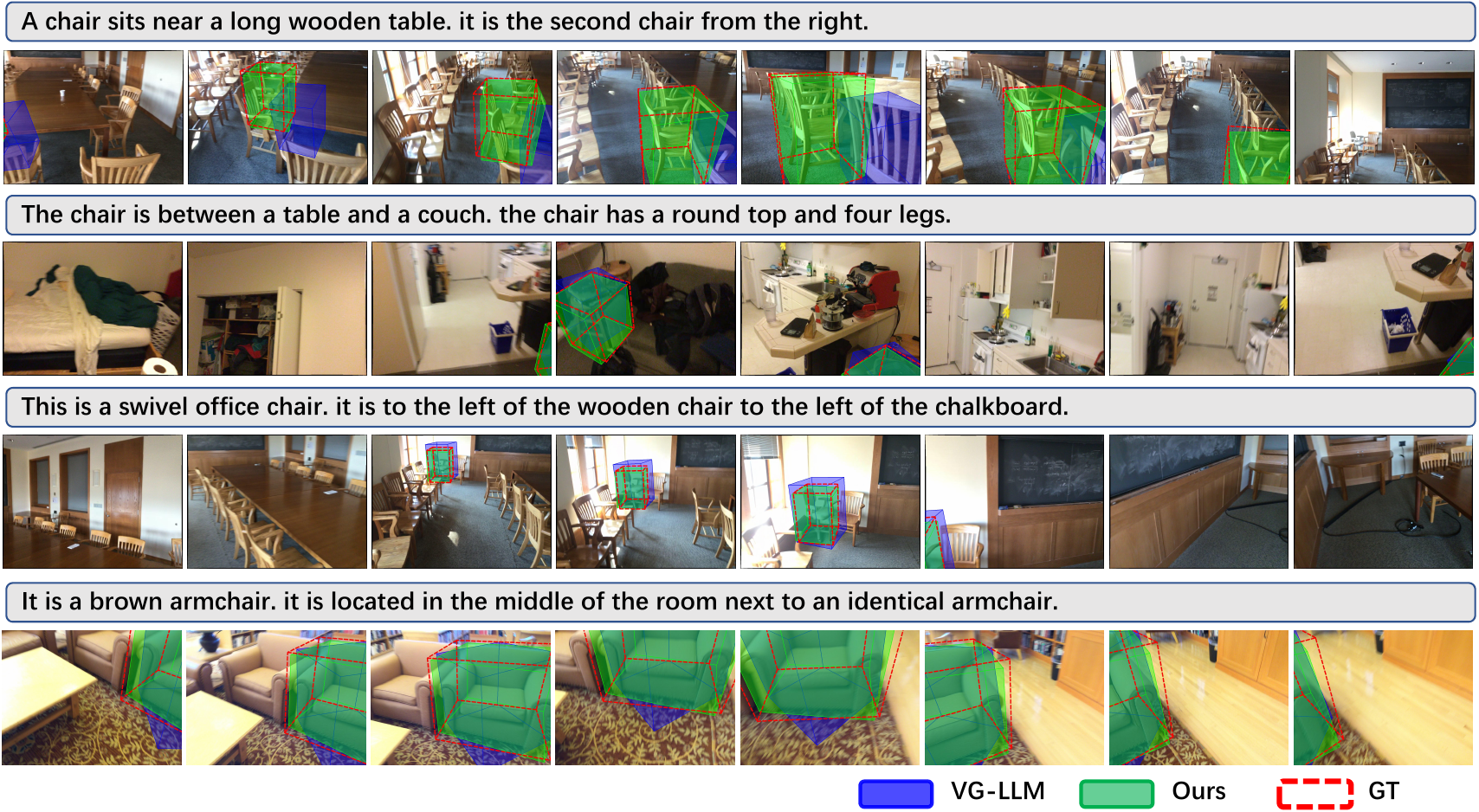}
\caption{
\textbf{Additional qualitative results on 3D visual grounding.}
Given a language query and multi-frame images, the model predicts the 3D bounding box of the described object.
Compared with VG-LLM, our method produces more accurate and stable localization results that better match the ground truth.
}
\label{fig:supp_grounding}
\end{figure*}

We provide additional qualitative examples of 3D visual grounding in Fig.~\ref{fig:supp_grounding}.
Compared with VG-LLM, our method produces more accurate and stable localization results that better align with the ground-truth annotations.
Although both methods successfully identify the target object, our predicted 3D boxes are more precise, suggesting that the proposed geometry-aligned pre-training paradigm leads to stronger geometric perception.

\subsection{3D Video Object Detection}

\begin{figure*}[htbp]
\centering
\includegraphics[width=\linewidth]{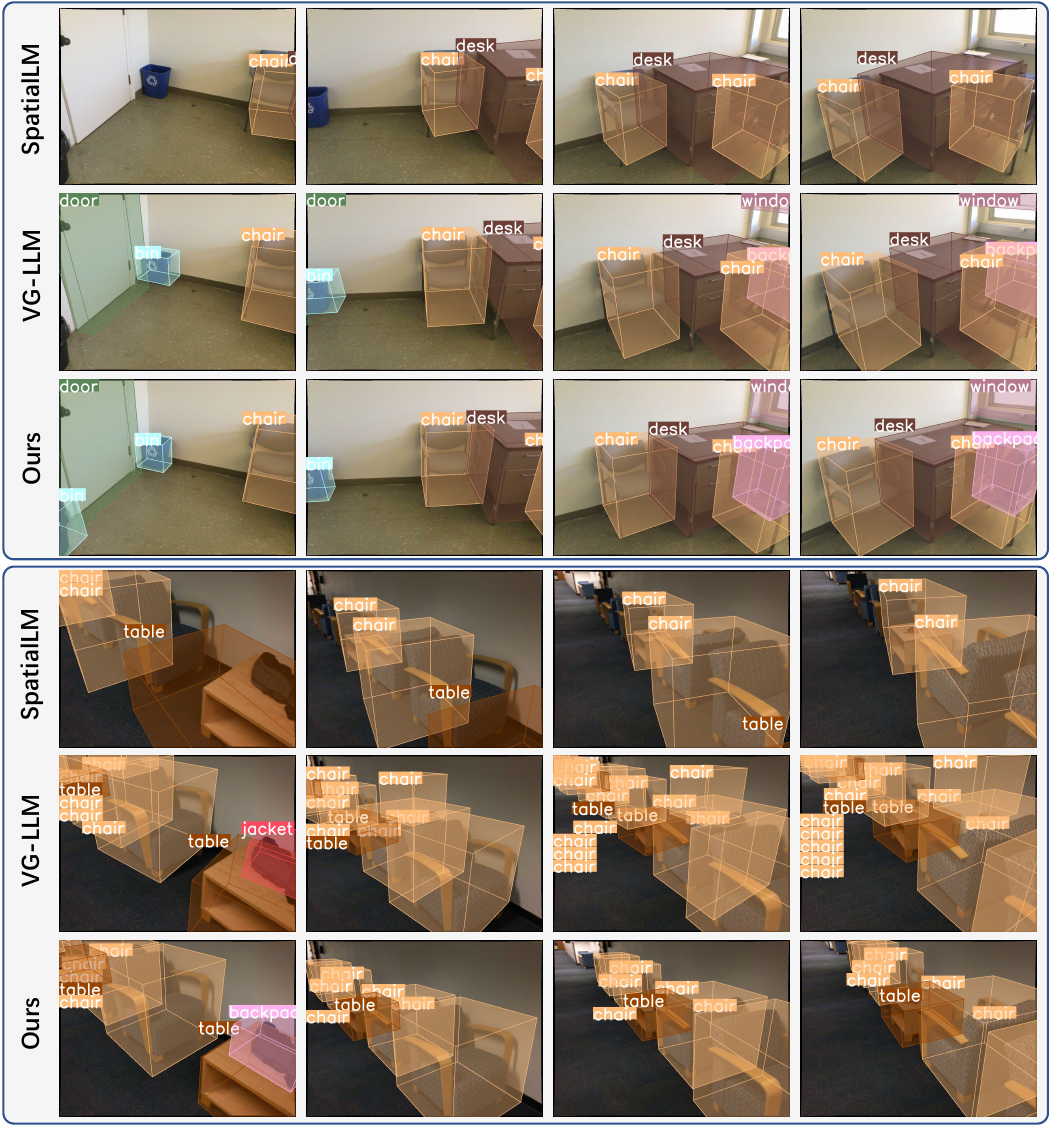}
\vspace{-3mm}
\caption{
\textbf{Additional qualitative results on 3D video object detection.}
Compared with SpatialLM~\cite{SpatialLM} and VG-LLM~\cite{zheng2025learning}, our method produces more accurate and consistent 3D bounding boxes for multiple objects in the scene, benefiting from improved geometric perception under the unified coordinate system.
}
\vspace{-10mm}
\label{fig:supp_detection}
\end{figure*}

Fig.~\ref{fig:supp_detection} presents additional qualitative results on 3D video object detection.
Under the unified coordinate system, the model predicts 3D bounding boxes for multiple objects in the scene.

Compared with SpatialLM~\cite{SpatialLM} and VG-LLM~\cite{zheng2025learning}, our method produces more consistent object localization and more accurate 3D box extents.
These results further demonstrate that the geometry-aligned pre-training enhances geometric perception, leading to more stable multi-object detection in complex scenes.

\section{Limitations and Future Work}
\label{sec:limitation}

\vspace{-2mm}

Although GAP-MLLM shows consistent improvements across multiple RGB-only 3D perception tasks, the current framework still has several limitations.

\noindent\textbf{Reliance on implicit geometric priors.}
Our method focuses on activating implicit geometric priors within MLLMs, rather than improving the reconstruction model itself.
In this work, we analyze and adopt a representative feed-forward geometric encoder to validate the effectiveness of the geometry-aligned pre-training paradigm.
Stronger geometric encoders may provide better priors and further improve downstream performance.

\noindent\textbf{Sparse supervision limits fine-grained reconstruction quality.}
Our experiments show that sparse joint supervision is sufficient to activate geometric perception and benefit downstream 3D perception.
However, compared with dense supervision, it is less effective for fine-grained geometry and semantics.
As a result, the reconstructed point clouds and semantic maps may remain coarse locally.
In particular, the predicted semantic maps inherit the abstract semantic behavior of MLLMs, which can lead to unclear boundaries.

\noindent\textbf{Future work.}
Future work may explore stronger geometric backbones, or directly train MLLMs to activate geometric perception without relying on external geometric priors.
Another promising direction is to design more effective pre-training strategies that reconstruct more precise pointmaps and semantic maps while further benefiting downstream 3D perception tasks, ultimately serving as a stronger foundation for 3D understanding and reasoning.

\end{document}